\RequirePackage{fix-cm}
\documentclass[smallcondensed]{svjour3}     
\smartqed  
\usepackage{graphicx,amsmath,amssymb,url,mathtools}
\usepackage{bbm}
\usepackage{algorithm}
\usepackage{algorithmic}
\usepackage{hyperref}
\usepackage{enumerate}
\usepackage{multirow}
\usepackage{subfigure}
 \usepackage{mathptmx}      
\def \cX {\mathcal{X}}
\def \cD {\mathcal{D}}
\def \cS {\mathcal{S}}

\def \Pr {\mathrm{P}}
\def \E  {\mathbf{E}}
\def \cH {\mathcal{H}}
\def \cC {\mathcal{C}}
\def \cF {\mathcal{F}}
\def \bal {\begin{align}}
\def \eal {\end{align}}

\def \bx {\mathbf{x}}
\def \bq {\mathbf{q}}

\def \err2 {\indicator{f_2(x^{1:2}_i) \neq y_i}}

\newcommand{\indicator}[1]{\mathbbm{1}_{\left[ {#1} \right] }}

\newcommand{\ev}[2]{\mathbf{E}_{#1} \left[ {#2} \right]}
\newcommand{\evc}[3]{\mathbf{E}_{#1} \left[ {#2} \mid {#3} \right]}

\newcommand{\prob}[2]{\mathrm{P}_{#1} \left[ {#2} \right]}
\newcommand{\sgn}[1]{\mathrm{sgn} \left[ #1 \right]}

\begin{document}

\title{Multi-Stage Classifier Design}



\author{Kirill Trapeznikov \and Venkatesh Saligrama \and David Casta$\tilde{\text{n}}\acute{\text{o}}$n}


\institute{K. Trapeznikov \at
              8 Saint MaryÕs Street, Boston, MA, 02215 \\
              Tel.: +1 (617) 353-2811 \\
              Fax:  +1 (617) 353-7337 \\
              \email{ktrap@bu.edu}   
           \and
           V. Saligrama \at
           8 Saint MaryÕs Street, Boston, MA, 02215 \\
              Tel.: +1 (617) 353-2811 \\
              Fax:  +1 (617) 353-7337 \\
              \email{srv@bu.edu}
              \and
           D. Castatnon \at
           8 Saint MaryÕs Street, Boston, MA, 02215 \\
              Tel.: +1 (617) 353-2811 \\
              Fax:  +1 (617) 353-7337 \\
              \email{dac@bu.edu}
}

\date{Received: date / Accepted: date}

\maketitle


\begin{abstract}
In many classification systems, sensing modalities have different acquisition costs. It is often {\it unnecessary} to use every modality to classify a majority of examples. We study a multi-stage system in a prediction time cost reduction setting, where the full data is available for training, but for a test example, measurements in a new modality can be acquired at each stage for an additional cost. We seek decision rules to reduce the average measurement acquisition cost. We formulate an empirical risk minimization problem (ERM) for a multi-stage reject classifier, wherein the stage $k$ classifier either classifies a sample using only the measurements acquired so far or rejects it to the next stage where more attributes can be acquired for a cost. To solve the ERM problem, we show that the optimal reject classifier at each stage is a combination of two binary classifiers, one biased towards positive examples and the other biased towards negative examples.  We use this parameterization to construct stage-by-stage global surrogate risk, develop an iterative algorithm in the boosting framework and present convergence and generalization results. We test our work on synthetic, medical and explosives detection datasets. Our results demonstrate that substantial cost reduction without a significant sacrifice in accuracy is achievable.
\end{abstract}


\section{Introduction}
In many applications including homeland security and medical diagnosis, decision systems are composed of an ordered sequence of stages. Each stage is associated with a sensor or a physical sensing modality. Typically, a less informative sensor is cheap (or fast) while a more informative sensor is either expensive or requires more time to acquire a measurement. In practice, a measurement budget (or throughput constraint) does not allow all the modalities to be used simultaneously in making decisions. The goal in these scenarios is to attempt to classify examples with low cost sensors and limit the number of examples for which more expensive or time consuming informative sensor is required.

\begin{figure}[htb!]
\centering
\includegraphics[width=.9 \linewidth]{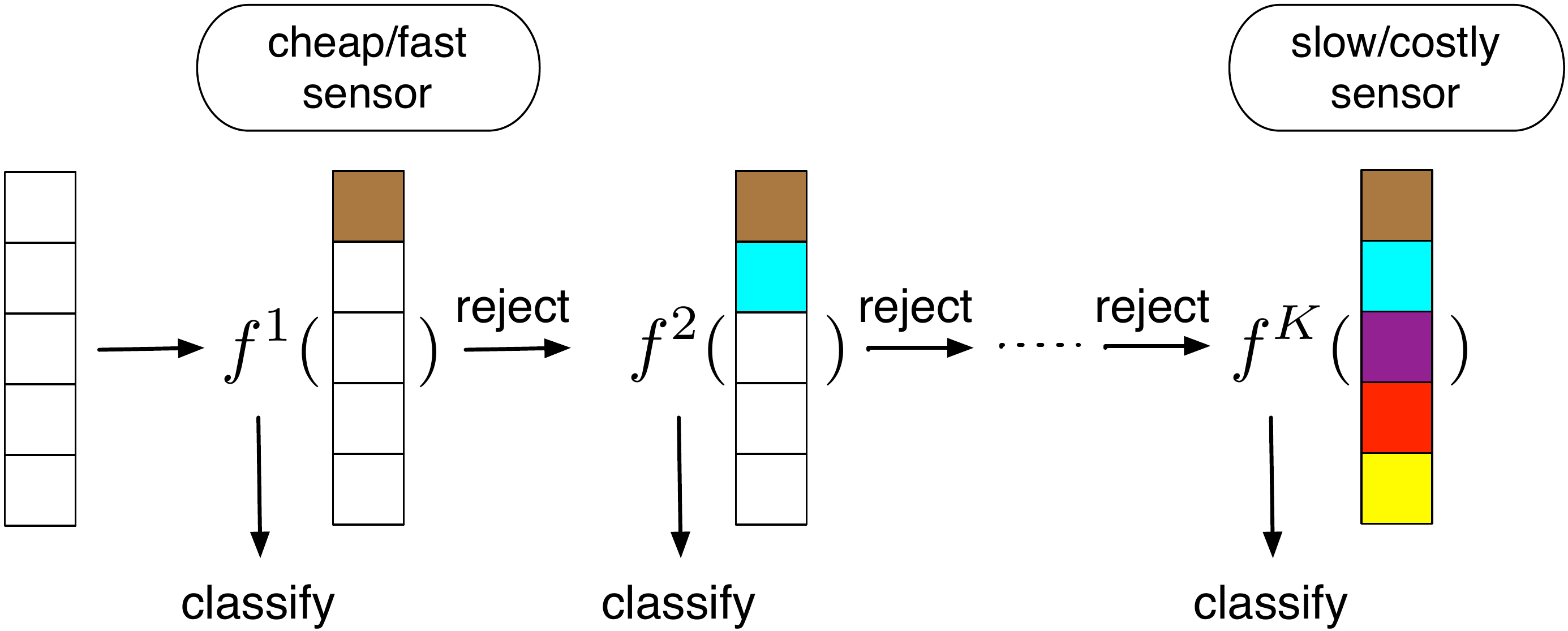}
\caption{ Multi-Stage System consists of $K$ stages. Each stage is a binary classifier with a reject option. The system incurs a penalty of $\delta_{k+1}$ at $k$th stage if it rejects to seek more measurements. The $k$th classifier only sees the first $k$ sensing modalities in making a decision.}
\label{fig:system}
\end{figure}

For example, in explosives detection, in the first stage, 
an infrared imager or a metal detector can be used with high throughput and low cost.  A second stage could be the the use of a slower, more expensive active millimeter wave (AMMW) scanner. The final third stage is a time consuming human inspection. In medical applications, first stages are typically non-invasive procedures (such as a physical exam) followed by more expensive tests (blood test, CT scan etc) and the final stages are invasive (surgical) procedures.

Many such examples share a common structure (see Fig. \ref{fig:system}), and we list some of its salient aspects below:

\paragraph {\bf (A)} \emph{Sensors \& Ordered Stages:} Each stage is associated with a new sensor measurement or a sensing modality. Multiple stages are an ordered sequence of sensors or sensor modalities with later stages corresponding to expensive or time-consuming measurements.
In many situations, there is often some flexibility in choosing a sensing modality from a collection of possible modalities. In these cases, the optimal choice of sensing actions also becomes an issue. While our methodology can be modified to account for this more general setting, we primarily consider a fixed order of stages and sensing modalities in this paper. This is justified on account of the fact that many of the situations we have come across consist of a handful of sensors or sensing modalities. Consequently, for these situations, the problem of choosing sensor ordering is not justified since one could by brute force enumerate and optimize over the different possibilities.

\paragraph {\bf (B)} \emph{Reject Classifiers:} Our sequential decision rules either attempt to fully classify an instance at each stage or "reject"  the instance on to the next stage for more measurements in case of ambiguity. For example, in explosives detection, a decision rule in the first stage, based on IR scan, would attempt to detect whether or not a person is a threat and identify the explosive type/location in case of a threat. If the person is identified as a threat at the first stage it is unnecessary (and indeed dangerous -- the explosive could be detonated) to seek more information. Similarly in medical diagnosis if a disease is diagnosed at an early stage, it makes sense to begin early treatment rather than waiting for more conclusive tests.

\paragraph {\bf (C)} \emph{Information vs. Computation}: Note that our setup can only use the partial measurements acquired up to a stage in making a decision. In other methods, such as detection cascades (\cite{viola01}), the full measurement and therefore all the information is available to every stage. Therefore, any region in the feature space can be carved out with more complex regions in the measurement space, or equivalently complex features can be extracted but with higher costs. In contrast, we have only partial measurements (or information) and so any feature or classifier that we employ has to be agnostic to unavailable measurements at that stage.

\begin{figure}[htb!]
\centering
\includegraphics[width=.9\linewidth]{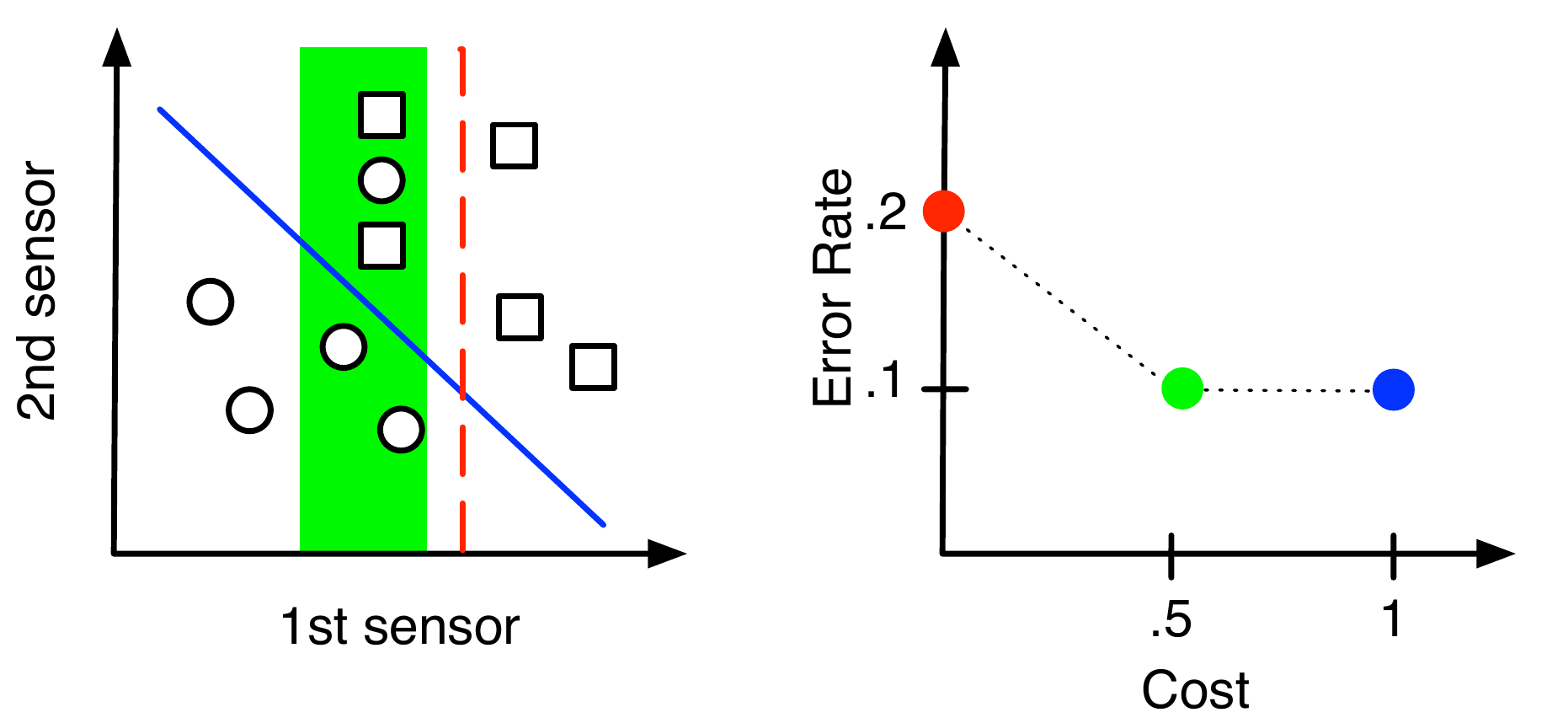}
\caption{(Advantage of a $2$ stage classifier: 10 samples, binary (squares, circles). The red line is the optimal decision when using only $1$st stage modality. The blue line is optimal if using both. (2nd stage) The curve is classification error vs. samples rejected (cost) The red point corresponds to classifying everything at stage $1$. The blue corresponds to rejecting everything and classifying using both modalities.(Stage 2) The green is a partial reject strategy. The samples outside the green region are classified using only the first modality, and samples inside the region are rejected to stage $2$ and are classified using both modalities. Note that blue and green have the same error, while the reject strategy (green) has to use $2$nd stage sensor only for ${1 \over 2}$ of examples, reducing the cost by a factor of $2$.}
\label{fig:toyexample}
\end{figure}

The two stage example in Fig. \ref{fig:toyexample} illustrates some of the advantages of our scheme over the alternative scheme that first acquires measurements from all the sensing modalities, which we refer to as the centralized classifier. A reject classifier utilizes the 2nd stage sensor only for a fraction of the data but achieves the same performance as the centralized classifier.

Our approach is based on the so called \emph{Prediction Time Cost Reduction approach} (\cite{kanani:2008}). Specifically, we assume a set of training examples in which measurements from all the sensors or sensing modalities as well as the ground truth labels are available. Our goal is to derive \emph{sequential reject classifiers} that reduces cost of measurement acquisition and error in the \emph{prediction (or testing) phase}.

We show that this sequential reject classifier problem can be formulated as an instance of a \emph{partially observable Markov Decision Process} {\bf (POMDP)} (\cite{kaelbling1998planning}) when the class-specific probability models for the different sensor measurements are known.
In this case the optimal sequential classifier can be cast as a solution to a Dynamic Program (DP). The DP solution is a sequence of \emph{stage-wise optimization} problems, where each stage problem is a combination of the cost from the current stage and the cost-to-go function that is carried on from later stages.

Nevertheless, class probability models are typically unknown; our scenarios produce high-dimensional sensor data (such as images). Consequently, unlike some of the conventional approaches (\cite{ji:2007}), where probability models are first estimated to solve POMDPs, we have to adopt a non-parametric \emph{discriminative learning} approach. We utilize the structure of the POMDP solution to empirically approximate the value of the cost-to-go function only at a discrete subset of the data-space. Next, instead of interpolating or parameterizing the cost-to-go function and learning it from data, we formulate an empirical discriminative objective that utilizes point-wise cost-to-go estimates evaluated on the training set and directly learn classifiers that minimize this objective. Using this decomposition, we formulate a novel \emph{multi-stage expected risk minimization (ERM) problem}. 
We solve this ERM problem at each stage by first factoring the cost function into classification and rejection decisions. Then we transform reject decisions into a binary classification problem. Specifically, we show that the optimal reject classifier at each stage is a combination of two binary classifiers, one biased towards positive examples and the other biased towards negative examples. The disagreement region of the two then defines the reject region.

We then approximate this empirical risk with a global surrogates. We present an iterative solution and demonstrate local convergence properties. The solution is obtained in a boosting framework. We then extend well-known margin-based generalization bounds (\cite{bartlett98}) to this multi-stage setting. 
We tested our methods on synthetic, medical and explosives datasets. Our results demonstrate an advantage of multistage classifier: cost reduction without a significant sacrifice in accuracy.

\subsection{Related Work}
\paragraph
{\bf Active Feature Acquisition (AFA):}
The subject of this paper is not new and has been studied in the Machine Learning community as early as \cite{MacKay_information-basedobjective}. Our work is closely related to the so called prediction time active feature acquisition (AFA) approach in the area of cost-sensitive learning. The goal there is to make sequential decisions of whether or not to acquire a new feature to improve prediction accuracy. A natural approach is to formalize a problem as an POMDP. \cite{ji:2007,kapoor:2009} model the decision process and infer feature dependencies while taking acquisition costs into account. \cite{Sheng06featurevalue,bilgic:2007,zubek:2002} study strategies for optimizing decision trees while minimizing acquisition costs. The construction is usually based on some purity metric such as entropy. \cite{kanani:2008} proposes a method that acquires an attribute if it increases an expected utility. However,  all these methods require estimating a probability likelihood that a certain feature value occurs given the features collected so far. While surrogates based on classifiers or regressors can be employed to estimate likelihoods, this approach requires discrete, binary or quantized attributes. In contrast, our problem domain deals with high dimensional measurements (images consisting of million of pixels), so we develop a discriminative learning approach and formulate a multi-stage empirical risk optimization problem to reduce measurement costs and misclassification errors. At each stage, we solve the reject classification problem by factorizing the cost function into classification and rejection decisions. We then embed the rejection decision into a binary classification problem.

\begin{figure}[htb!]
\begin{center}
\subfigure[]{\includegraphics[height=.3\linewidth]{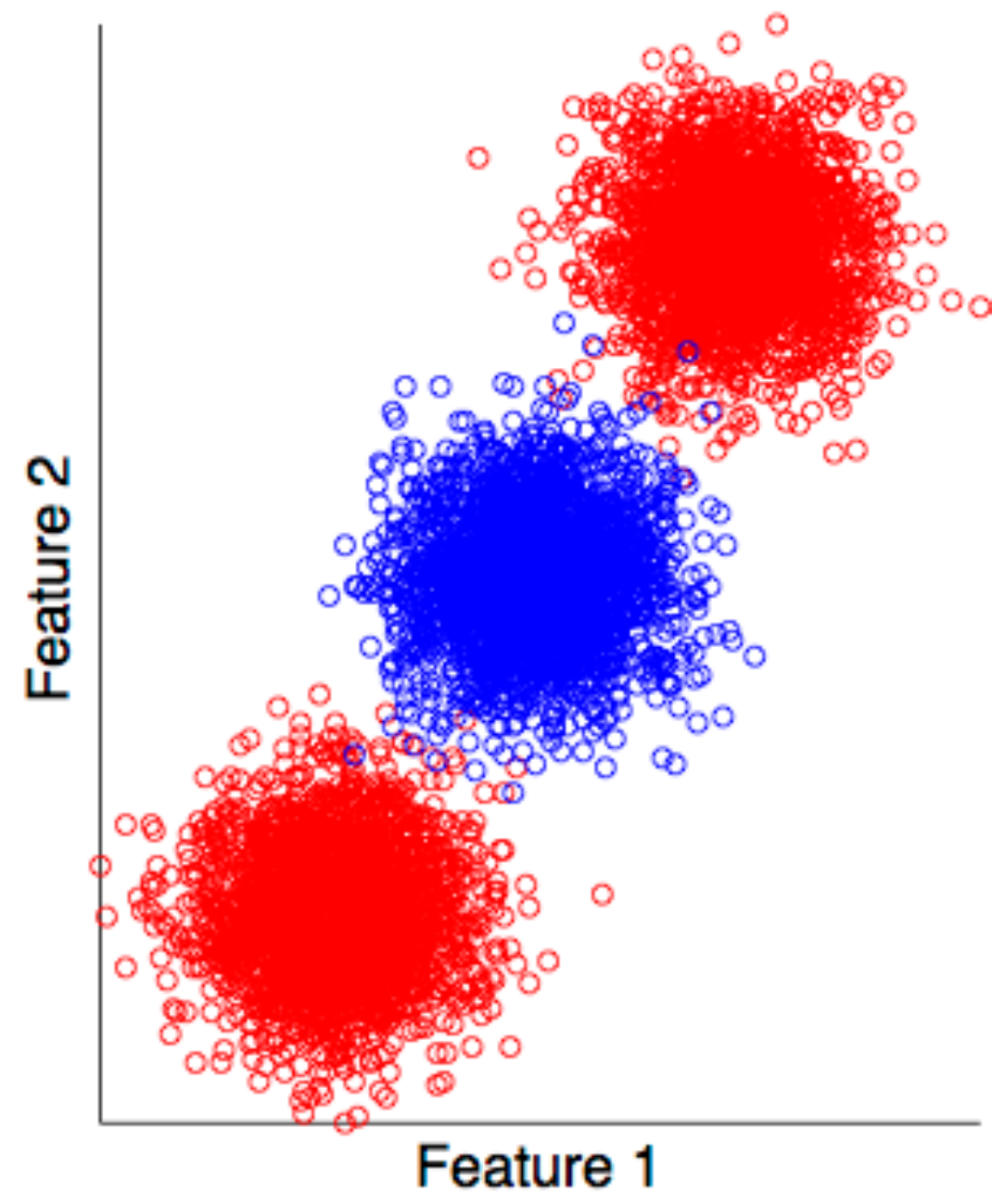}}
\subfigure[]{\includegraphics[height=.3 \linewidth]{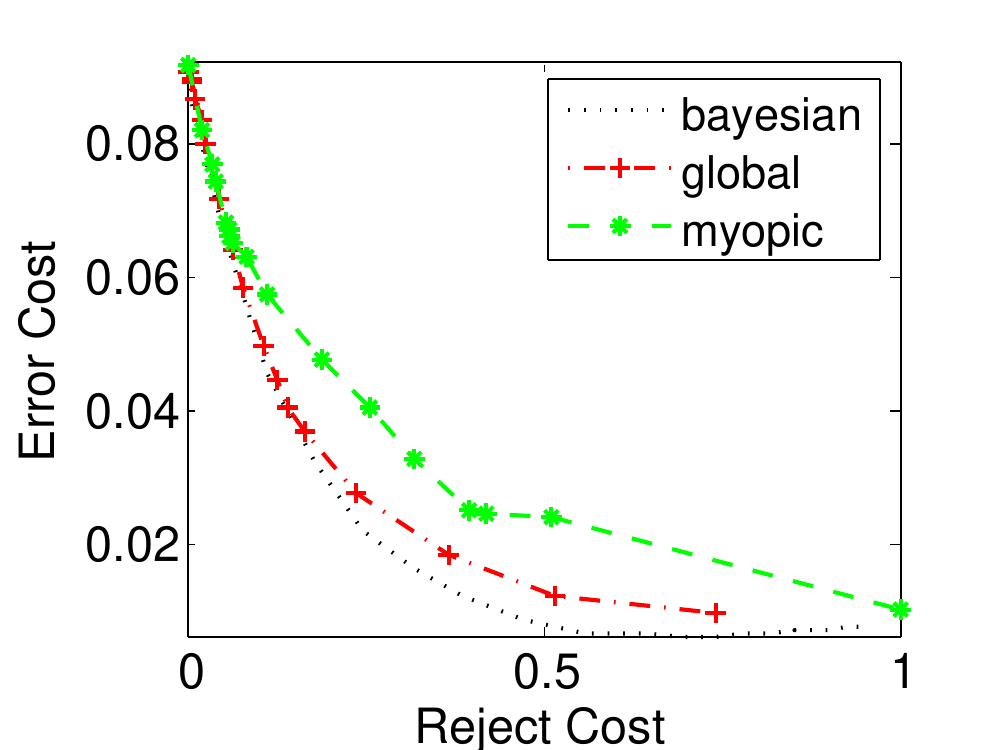}}
\caption{ (a) Gaussian Mixture (binary). (b) Error rate vs reject rate on complementary measurements. 1st stage uses only dim 1. 2nd stage uses only dim. 2. Myopic strategy (green) is thresholding the margin of the classifier, our method is global surrogate; Bayesian classifier (best performance). Thresholding the margin performs significantly worse than our method.}
\label{fig:gauss_mix}
\end{center}
\end{figure}

\paragraph
{\bf Single Stage Reject Classifiers:}
Our paper is also closely related to the topic of reject classifiers, which has also been investigated. However, in the literature reject classifiers have been primarily considered in a single stage scenario. In the Bayesian framework, \cite{chow70} introduced Chow's rule for classification. It states that given an observation $x$ and a reject cost $\delta$ and $J$ classes, reject $x$ if the maximum of the posteriors for each class is less than the reject cost: $\max_{k=1 .. J} \Pr(y=j|x) < \delta$. In the context of machine learning, the posterior distributions are not known, and a decision rule is estimated directly. One popular approach is to reject examples with a small margin. Specifically, in the context of support vector machine classifiers, \cite{yuan03,barlett08,diaz09,grandvalet:2008}, define a reject region to lie within a small distance (margin) to the separating hyperplane and embed this in the hinge loss of the SVM formulation. \cite{el-yaniv11} proposes a reject criteria motivated by active learning but its implementation turns out to be computationally impractical. In contrast, we consider multiple stages of reject classifiers. We assume an error prone second stage which occurs in such fields as threat detection and medical imaging. In this scenario, rejecting in the margin is not always meaningful. Fig. \ref{fig:gauss_mix} illustrates that thresholding the margin to reject can lead to significant degradation. This usually happens when stage measurements are complimentary; then examples within a small margin of the 1st stage boundary may not be meaningful to reject. Multiple stages of margin based reject classifiers have been considered by \cite{liu08} using SVMs in image classification. The method does not take into account the cost of later stages and is similar to the myopic method that we compare in the Experiments section.

\paragraph
{\bf Detection Cascades:}
Our multi-stage sequential reject classifiers bears close resemblance to detection cascades. There is much literature on cascade design (see \cite{zhang:2010,chen:2012} and references therein) but most cascades roughly follow the set-up introduced by \cite{viola01} to reduce computation cost during classification. At each stage in a cascade, there is a binary classifier with a very high detection rate and a mediocre false alarm rate. Each stage makes a partial decision; it either detects an instance as negative or passes it on to the next stage. Only the last stage in the cascade makes a full decision, namely, whether the example belongs to a positive or negative class.

There are several fundamental differences between detection cascades and the multi-stage reject classifiers (MSRC). A key difference is the system architecture. Detection cascades are primarily concerned with binary classification problems. They make partial decisions, delaying a positive decision until the final stage. In contrast, MSRCs can make full classification decisions at any stage. Conceptually, this distinction requires a fundamentally new approach; detection cascades work because their focus is on unbalanced problems with few positives and a large number of negatives; and so the goal at each stage is to admit large false positives with negligible missed detections. Consequently, each stage can be associated with a binary classification problem that is acutely sensitive to missed detections. In contrast, our scheme at each stage is a composite scheme composed of a classifier as well as a rejection decision. The rejection decision is itself a binary classification problem. In practice, MSRCs arise in important areas such as medical diagnosis and explosives detection as we argued in Sec~1, item {\bf (B)}. As a performance metric detection cascades tradeoff missed detections at the final stage with average computation. MSRC's tradeoff average misclassification errors against number of examples that reached later stages (i.e. required more sensors or sensing modalities). For these reasons it is difficult to directly compare algorithms developed for MSRCs to those developed for detection cascades. Nevertheless, our goals and resulting algorithms are similar to some of the issues that arise in cascade design (see \cite{chen:2012} and references therein), namely, perform a joint optimization for all the stages in a cascade given a cost structure for different features.

\paragraph {\bf Other Cost Sensitive Methods:}
Network intrusion detection systems (IDS) is an area where sequential decision systems have been explored. (see \cite{Fan2000,Lee2002,Cordella2007}). In IDS, features have different computation costs. For each cost level, a ruleset is learned. The goal is to use as many low cost rules as possible. In a related set-up, \cite{Fan2002,Wang2003} consider a more general ensemble of base classifiers and explore how to minimize the ensemble size without sacrificing performance.  In the test phase, for a sample, another classifier is added to the ensemble if the confidence of the current classification low. Here, similar to detection cascades, the goal is to reduce computation time. As we described in Sec~1, item {\bf (C)}, the important distinction is that, in our setting, a decision is based only on the partial information acquired up to a stage. In a computation driven method, a stage (or base classifier) decides using a feature computed from the full measurement vector.

\section{Problem Statement}
Let $(\bx,y) \in \cX \times \{1,2, \ldots C\}$ be distributed according to an unknown distribution $\cD$. A data point has $K$ features, $\bx=\{x_1,x_2, \ldots, x_K\}$, and belongs to one of $C$ classes indicated by its label $y$. A $k$th feature is extracted from a measurement acquired at $k$th stage. We define a truncated feature vector at $k$th stage: $\bx^k = \{x_1, x_2, \ldots x_k\}$. Let $\cX^k$ be the space of the first $k$ features such that $\bx^k \in \cX^k$.

The system has $K$ stages, the order of the stages is fixed, and $k$th stage acquires a $k$th measurement. At each stage, $k$, there is a decison with a reject option, $f^k$.  It can either classify an example, $f^k(x^{k}): \cX^k \to \{1,2, \ldots, C\}$, or delay the decision until the next stage, $f^k(x^{k})=r$ and incur a penalty of $\delta^{k+1}$. Here, $r$ indicates the "reject" decision. $f^k$ has to make a decision using only the first $k$ sensing modalities. The last stage $K$ is terminal, a standard classifier. Define the system risk to be,
\begin{align}
R(f^1,\ldots, f^K,x,y) =
 \sum_{k=1}^K  S^k(\bx^k) R_k(f^k,\bx^k,y) \label{eq:single}
 \end{align}
Here, $R_k$ is the cost of classifying at $k$th stage, and $S^k(\bx^k) \in \{0,1\}$ is the binary state variable indicating whether $x$ has been rejected up to $k$th stage.
\begin{align}
R_k(\bx^k,y,f^k)&=
 \begin{cases}
 \delta^{k+1}, & f^k(\bx^k)=r \\
 1, & f^k(\bx^k) \neq y~\land~f^k(\bx^k) \neq r
\end{cases} \nonumber 
\end{align}
If $x$ is active and is misclassified, the penalty is $1$ \footnote{To simplify our discussion, we consider equal error penalties. However, our approach can be easily extended to unbalanced error penalties as we will demonstrate in the experiments section}. If it is rejected then the system incurs a penalty of $\delta^{k+1}$, and the state variable for that example remains at $1$.
\begin{align}
S^{k+1}(\bx^{k+1})&=
\begin{cases} S^{k}(\bx^{k}),& f^k(\bx^k) = r \\
0, & \mbox{else}
\end{cases},~S^1=1 \label{eq:sk}
\end{align}

\subsection{Bayesian Setting}
In this section, we will digress from the discriminative setting and analyze the problem under the assumption that the underlying distribution $\cD$ is known. In doing so, we hope to discover some fundamental structure that will simplify our empirical risk formulation in the next section.

If $\cD$ is known the problem reduces to an POMDP, and the optimal strategy is to minimize the expected risk,
\begin{align}
\min_{f^1, \ldots, f^K} \ev{\cD}{R(f^1, \ldots, f^K,\bx^k,y)} \label{eq:uncondrisk}
\end{align}
If we allow arbitrary decision functions then we can equivalently minimize conditional risk,
\begin{align}
\min_{f^1, \ldots, f^K} \evc{}{ R(f^1, \ldots, f^K,\bx^k,y)}{\bx} \label{eq:condrisk}
\end{align}
This problem---by appealing to dynamic programming---remarkably reduces to a single stage optimization problem for a modified risk function. To see this, we denote the cost-to-go,
\begin{align}
\tilde \delta^k(\bx^k)= \delta^{k+1} +
\min_{f^{k+1} \ldots f^K} \evc{}{\sum_{t=k+1}^K S^t(\bx^t) R_t(f^t,\bx^t,y)}{\bx^k, S^k(\bx^k)=1}  \nonumber
\end{align}
and the modified risk functional, 
\begin{align}
\tilde R_k(\bx^k,y,f^k,\tilde \delta^k)&=
 \begin{cases}
 \tilde \delta^k(\bx^k),&f^k(\bx^k)=r  \\
 1,&f^k(\bx^k) \neq y~\land~f^k(\bx^k) \neq r
\end{cases} \nonumber 
\end{align}
and prove the following theorem,
\newtheorem{thm1}{Theorem}
\begin{thm1}
The optimal solution $f^1,f^2, \ldots f^K$ to the multi-stage risk in Eq. \ref{eq:condrisk} decomposes to single stage optimization, 
\begin{align}
f^k=\arg \min_{f} \evc{}{\tilde R_k(\bx^k,y,f,\tilde \delta^k)}{\bx^k} \label{eq:trkopt}
\end{align}
and the solution is:
\begin{align}
&f^k(\bx^k)=
\begin{dcases}
\hat y, & \bar \Pr(\bx^k) > {1 - \tilde  \delta^k(\bx^k)}  \\
\mbox{reject}, & \bar \Pr( \bx^k) \leq 1 -{\tilde  \delta^k(\bx^k)} 
\end{dcases} \\
&\hat y =\arg\max_{j} \Pr(y= j \mid \bx^k),~ \bar \Pr(\bx^k)= \max_{j} \Pr(y=j \mid \bx^k) \nonumber
\end{align}
\end{thm1}
\begin{proof}
To simplify our derivations, we assume uniform class prior probability: $\prob{y}{y=\hat y}=\frac{1}{c},~ \hat y=1, \ldots, C$. However, our results can be easily modified to account for a non-uniform prior.  The expected conditional risk can be solved optimally by a dynamic program, where a DP recursion is,
\begin{align}
J_{K}(\bx^K,S^K)=\min_{f^K} \evc{y}{S^K(\bx^K) R_k(y,\bx^K,f^K)}{\bx^K} \\
J_{k}(\bx^k,S^k)= \min_{f^k} \left \{  \evc{y}{S^k(\bx^k) R_k(y,\bx^k,f^k)}{\bx^k}+  
\evc{\bx^{k+1} \ldots \bx^K}{J_{k+1}(\bx^{k+1},S^{k+1})}{\bx^k} \right \}
\end{align}
Consider $k$th stage minimization, $f^k$ can take $C+1$ possible values $\{1,2, \ldots C,~r\}$ and $J_{k}(\bx^k,S^k)$ can be recast as an conditional expected risk minimization,
{ \begin{align}
J_{k}(\bx^k,S^k=1)= 
\min_{f^k} \left \{ \underbrace{ \prob{y}{y \neq \hat y \mid \bx^k}}_{f^k(\bx^k)=\hat y}, \underbrace{\delta^k + \evc{\bx^{k+1} \ldots \bx^K}{J_{k+1}(\bx^{k+1},1)}{\bx^k}}_{f^k(\bx^k)=r} \right \} \label{eq:jk}
\end{align}}
Define,
\[\tilde \delta(x^k)=\delta^{k+1} + \evc{\bx^{k+1} \ldots x^K}{J_{k+1}(\bx^{k+1},S^{k+1}=1)}{\bx^k}\]
and rewrite the conditional risk in \ref{eq:jk},
\begin{align}
f^k = \arg \min_{f} \left \{ \underbrace{ 1- \prob{y}{y = \hat y \mid \bx^k}}_{f(\bx^k)=\hat y}, \underbrace{\tilde \delta^k(\bx^k)}_{f(\bx^k)=r} \right \}
\end{align}
Reject is the optimal decision if,
\begin{align}
\min_{\hat y} \left \{ 1- \prob{y}{y = \hat y \mid \bx^k} \right \} \geq \tilde \delta^k(\bx^k)  \implies  \max_{\hat y} \left \{ \prob{y}{y = \hat y \mid \bx^k} \right \} \leq 1- \tilde \delta^k(\bx^k) 
\end{align}
If reject is not the optimal strategy then a class is chosen to maximize the posterior probability: 
\begin{align}
f^k(\bx^k)= \arg \max_{\hat y \in \{1, \ldots, c\}} \left \{ \prob{y}{y = \hat y \mid \bx^k} \right \}
\end{align}
which is exactly our claim.
\qed
\end{proof}

The main implication of this result is that if the cost-to-go function $\tilde \delta^k(\bx^k)$ is known then the risk $\tilde R_k(\cdot)$ is only a function of the current stage decision $f^k$. Therefore, we can ignore all of the other stages and minimize a single stage risk. Effectively, we decomposed the multi-stage problem in Eq. \ref{eq:condrisk} into a stage-wise optimization in Eq. \ref{eq:trkopt}.

Note that the modified risk functional, $\tilde R_k$, is remarkably similar to $R_k$ except that the modified reject cost $\tilde \delta^k(\bx^k)$ replaces the constant stage cost $\delta^k$. Also, consider the range for which $\delta^k(\bx^k)$ is meaningful. If we have $C$ classes then a random guessing strategy would incur an average risk of $1-\frac{1}{C}$. Therefore the risk for rejecting,
$\tilde \delta^k(\bx^k) \leq 1-\frac{1}{C}$
in order to be a meaningful option. The work in \cite{chow70} contains a detailed analysis of single stage reject classifier in a Bayesian setting.
\begin{figure}[htb!]
\centering
\includegraphics[width=.6\linewidth]{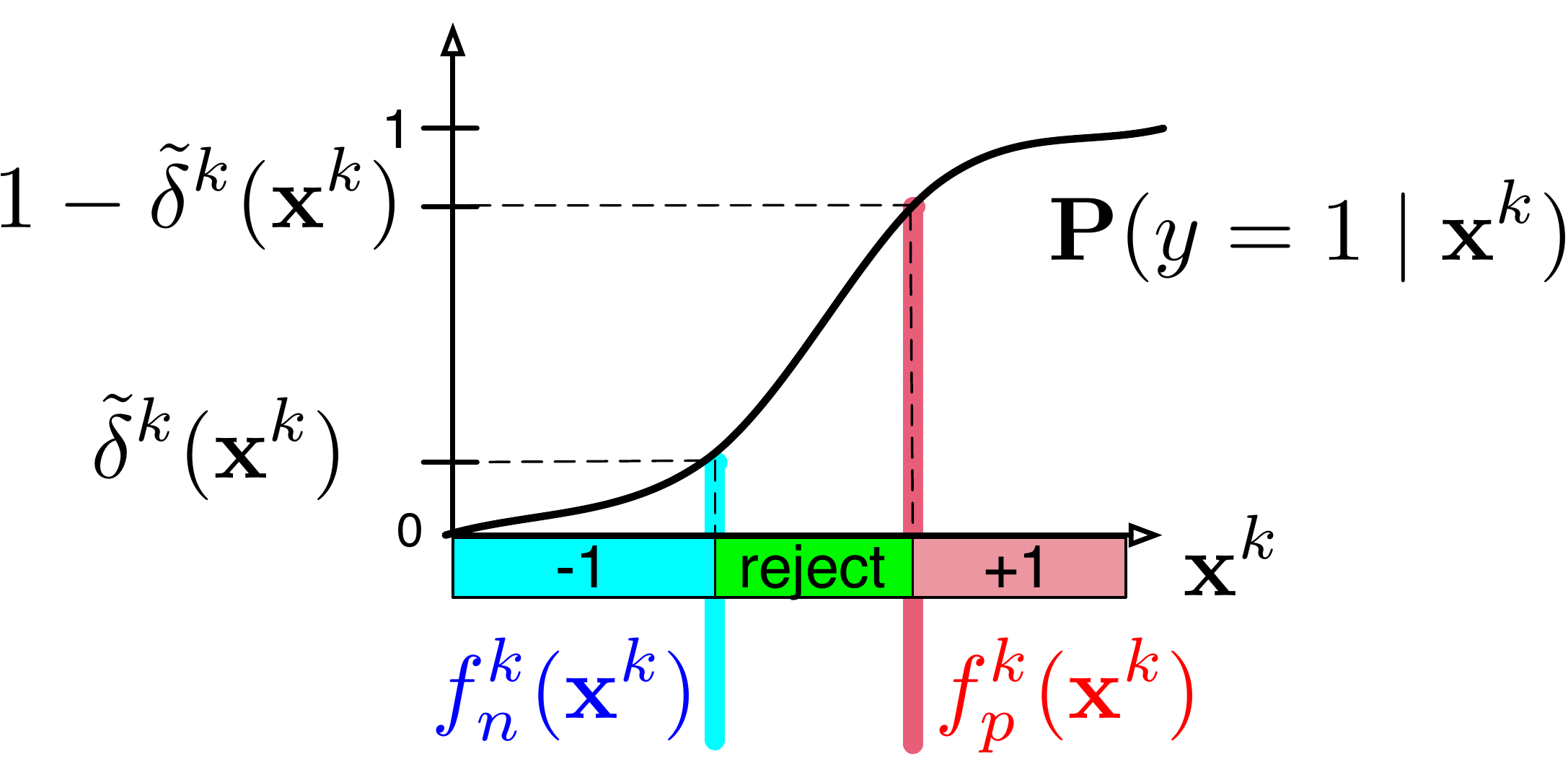}
\caption{Optimal Reject Region can be expressed as the disagreement region of two binary classifiers ($f_n$ and $f_p$)}.
\label{fig:region}
\end{figure}
\paragraph{Reject Classifier As Two Binary Decisions:} Consider a stage $k$ classifier with a reject option from Theorem 1 in a binary classification setting, $y \in \{-1,+1\}$.
\begin{align}
f^k(\bx^k)=
\begin{cases}
+1, &  \Pr(y=1 \mid \bx^k) > {1 - \tilde  \delta^k(\bx^k)}  \\
-1, &  \Pr(y=1 \mid \bx^k) < \tilde  \delta^k(\bx^k)  \\
\mbox{reject}, &  \delta^k(\bx^k) \leq \Pr(y=1 \mid \bx^k) \leq 1 -{\tilde  \delta^k(\bx^k)} 
\end{cases} \label{eq:optbayesbin}
\end{align}
It is clear from the expression that we can express the decision regions in terms of two binary classifiers $f_n$ and $f_p$. Observe that for a given reject cost $\tilde \delta^k(\bx^k)$, the reject region is an intersection of two binary decision regions. To this end we further modify the risk function in terms of agreement and disagreement regions of the two classifiers, $f_n,\,f_p$, namely,
\begin{align}
L_k(\bx^k,y,f_n,f_p,\tilde \delta^k)=
 \begin{cases}
  \tilde \delta(\bx^k),& f_n(\bx^k) \not =f_p(\bx^k) \\
   1, &  f_n(\bx^k) = f_p(\bx^k) \land f_p(\bx^k) \neq y
\end{cases}
\end{align}
Note that the above loss function is symmetric between $f_n$ and $f_p$ and so any optimal solution can be interchanged. Nevertheless, we claim:
\newtheorem{thm2}[thm1]{Theorem}
\begin{thm2}
Suppose $f_n$ and $f_p$ are two binary classifiers that minimize $\E \left [ L_k(\bx^k,y,f_n,f_p,\tilde \delta^k) \mid \bx^k \right ]$ over all binary classifiers $f_n$ and $f_p$. Then following resulting reject classifier:
\begin{align}
f^k(\bx^k)=  \begin{cases}
     f_p(\bx^k), & f_n(\bx^k)=f_p(\bx^k) \\
        \mbox{reject}, & f_n(\bx^k) \neq f_p(\bx^k)
        \end{cases} \label{optbayes_dec}
\end{align}
is the minimizer for $\E \left [ \tilde R_k(\bx^k,y,f,\tilde \delta^k) \mid \bx^k \right ]$ in Theorem 1 and the $k$th stage minimizer in Eq.~\ref{eq:uncondrisk}. 

\end{thm2}
\begin{proof}
For a given $\bx^k$ and $\tilde \delta(\bx^k)$,
\begin{align}
&\min_{f} \evc{y}{\tilde R_k(\bx^k,y,f,\tilde \delta^k)}{\bx^k} = &\min_{f} \left \{ \underbrace{ \prob{y}{y=-1 \mid \bx^k}}_{f=+1} , \underbrace{ \prob{y}{y=+1 \mid \bx^k}}_{f=-1}, \underbrace{\tilde \delta(\bx^k)}_{f=\mbox{\small reject}}  \right \} \nonumber \\
&\min_{f_p,f_n} \evc{y}{L_k(\bx^k,y,f_p,f_n,\tilde \delta^k)}{\bx^k}= &\min_{f_p,f_n} \left \{ \underbrace{ \prob{y}{y=-1 \mid \bx^k}}_{f_p=+1,f_n=+1} , \underbrace{\prob{y}{y=+1 \mid \bx^k}}_{f_p=-1,f_n=-1}, \underbrace{\tilde \delta^k(\bx^k)}_{f_p \neq f_n} \right \} \nonumber
\end{align}
By inspection, the decomposition in \ref{optbayes_dec} is the optimal bayesian classifier minimizing $\evc{y}{\tilde R_k(\bx^k,y,f,\tilde \delta^k)}{\bx^k}$
\qed
\end{proof}
We refer to Fig \ref{fig:region} for an illustration. We can express the new loss compactly as follows:
\begin{align}
L_k(\bx^k,y,f_p,f_n,\tilde \delta^k)= \indicator{f_p(\bx^k) \neq y}\indicator{f_n(\bx^k) \neq y} + \tilde \delta^k(\bx^k) \indicator{f_p(\bx^k) \neq f_n(\bx^k)} \label{eq:Lk}
\end{align}
Note that in arriving at this expression we have used: $\indicator{a \neq c} \indicator{a = b} = \indicator{a \neq c} \indicator{b \neq c}$. 

In summary, in this section, we derive the optimal POMDP solution and decouple a multi-stage risk to single stage optimization. Then, for the binary classification setting, we derive an optimal representation for a reject region classifier in terms of two biased binary decisions:
\[
\min_{f^k} \E [ R(\bx,y, \ldots, f^k, \ldots ] \rightarrow \min_{f^k} \E [ \tilde R_k(\bx^k,y, f^k,\tilde \delta^k) ]  \rightarrow \min_{f_p^k,f_n^k} \E [ L_k(\bx^k,y, f_p^k,f_n^k, \tilde \delta^k) ]
\]

\subsection{Stage-wise Empirical Minimization}
In this section, we assume that the probability model $\cD$ is no longer known and cannot be estimated due to high-dimensionality of the data.
Instead, our task is to find multi-stage decision rules based on a given training set: $(\bx_1,y_1), (\bx_2,y_2), \ldots ,(\bx_N,y_N)$. Here, we consider binary classification setting: $y_i \in \{+1,-1\}$. 

We will take advantage of the stage-wise decomposition of the POMDP solution in Theorem 1 and parametrization of reject region in Theorem 2 to formulate an empirical version of the stage risk $L_k(\cdot)$ in Eq. \ref{eq:Lk}. However, this requires the knowledge of the cost-to-go, $\tilde \delta^k: \cX^k \to \mathbbm R$. Instead of trying to learn this complex function, we will define a point-wise empirical estimate of the cost-to-go on the training data:
\[ \tilde \delta^k(\bx^k_i) \to \tilde \delta^k_i~,i=1,2, \ldots N\]
and use it to learn the decision boundaries directly. 
 
Note that by definition, $\tilde \delta^k(\bx^k_i)$ is a only function of $f^{k+1}, \ldots, f^K$. So the cost-to-go estimate is conveniently defined by the recursion,
\begin{align}
\tilde \delta^{k-1}_i=L_k(\bx^k_i,y_i,f_p^k,f_n^k,\tilde \delta_i^k) + \delta^{k},~ \forall i  \label{eq:deltak}
\end{align} 

Now, we can form the empirical version of the risk in Eq \ref{eq:trkopt} and optimize for a solution at stage $k$ over some family of functions, $\cF^k$.
\begin{align}
\{f_p^k(\bx^k),f_n^k(\bx^k)\} = \arg\min_{f_p,f_n \in \{ \cF^k\}^2} {1 \over N} \sum_{i=1}^N S^k_i L_k(\bx^k_i,y_i,f_p,f_n,\tilde \delta_i^k) \label{eq:emprisk}
\end{align}
Observe that, as in standard setting, we need to constrain the class of decision rules $f^k_p,f^k_n \in \cF_k \times \cF_K$ here. This is because with no constraints the minimum risk is equal to zero and can be achieved in the first stage itself.

Note, our stage-wise decomposition significantly simplifies the ERM. The objective in Eq. \ref{eq:emprisk} is only a function of $f_p^k,f_n^k$ given $\tilde \delta^k_i$ and the state $S^k_i$. To minimize an empirical version of a multi-stage risk in Eq. \ref{eq:uncondrisk} is much more difficult due to stage interdependencies.

Given $\delta^k_i$ and all the stages but the $k$th, we can solve \ref{eq:emprisk} by iterating between $f^k_p$ and $f^k_n$. To solve for $f^k_p$, we fix $f^k_n$ and minimize a weighted error 
\begin{align}
f^k_p=\arg \min_{f \in \cF^k} \sum_{i=1}^N w_i \indicator{f(\bx^k_i) \neq y_i} \label{subproblem},~w_i=S_i^k \left [ \indicator{f^k_n(\bx^k_i) \neq y_i}+\tilde \delta^k_i- 2 \indicator{f^k_n(\bx^k_i) \neq y_i} \tilde \delta^k_i \right ] 
\end{align}
We can solve for $f_n$ in the same fashion by fixing $f_p$,
\begin{align}
f^k_n=\arg \min_{f \in \cF^k} \sum_{i=1}^N w_i \indicator{f(\bx^k_i) \neq y_i},~w_i=S_i^k \left [ \indicator{f^k_p(\bx^k_i) \neq y_i}+\tilde \delta^k_i- 2 \indicator{f^k_p(\bx^k_i) \neq y_i} \tilde \delta^k_i \right ] 
\end{align}
To derive these expressions from \ref{eq:emprisk}, we used another identity for any binary variables $a,b,c$
\begin{align}
\indicator{a \neq b}= \indicator{a \neq c} + \indicator {b \neq c} - 2 \indicator {a \neq c} \indicator {b \neq c} 
\end{align}

\section{Algorithm}
Minimizing the indicator loss is a hard problem. Instead, we take the usual ERM (empirical risk minimization) (\cite{friedman2001elements}) approach and replace it with a surrogate. We introduce an algorithm in the boosting framework based on the analysis from the previous section. Boosting is just one of our many possible machine learning approaches that can be used to solve it. We use boosting because it is easy to implement and is known to have good performance.

Boosting is a way to combine simple classifiers to form a strong classifier. We are given a set of such weak classifiers  $\cH=\{h_1(\bx), h_2(\bx) \ldots h_M(\bx) \},~h_j(\bx) \in \{-1,+1\}$. The strong classifier is the linear combination: 
\[ F(\bx)=\sgn{\sum_{h_j \in \cH} q_{j} h_j(\bx)}\]
This set of weak classifiers need not be finite. Also, denote $\cH_k \subset \cH$ as a subset of weak classifiers that operate only on the first $k$ measurements of $x$. $h_j(\bx)=h_j(\bx^{k})~if~ h_j \in \cH_k$ .

\paragraph{Global Surrogate:} In our algorithm, we use the sigmoid loss function $\mathbf C(z)={1 \over 1+exp(z)} $ to approximate the indicator. Similar sigmoid based losses have been used in boosting before (\cite{vasc09}). Each subproblem (\ref{subproblem}) reduces to boosting a weighted loss

To solve for stage $k$, we keep the rest of the stages constant. To find $f^k_p=\sum q_{j} h_j(\bx)$, we fix $f_n^k$ and solve:
\begin{align}
f^k_p=\arg \min_{q_1, q_2, \ldots} \sum_{i=1}^N w_i \mathbf C \left (y_i \sum_{h_j \in \cH^k} q_j h_j(\bx_i)\right ) \label{eq:fp}
\end{align}
Note that the weights $w_i$, state variables $S^k_i$ and cost-to-go $\tilde \delta^k_i$ are also expressed in terms of the $\mathbf C(z)$ instead of $\indicator{z}$:
\begin{eqnarray}
w_i=S_i^{k} \left [ \mathbf C(y f^k_n(\bx_i))+\tilde \delta^k_i- 2 \mathbf C(y f^k_n(\bx_i)) \tilde \delta^k_i \right ]
\end{eqnarray}
To solve for $f_n^k$, we solve the same problem but keep $f_p^k$ constant instead:
\begin{align}
f^k_n=\arg \min_{q_1, q_2, \ldots} \sum_{i=1}^N  w_i \mathbf C \left (y_i \sum_{h_j \in \cH^k} q_j h_j(\bx_i)\right ) \label{eq:fn}  \\ 
w_i=S_i^{k} \left [ \mathbf C(y f^k_p(\bx_i))+\tilde \delta^k_i- 2 \mathbf C(y f^k_p(\bx_i)) \tilde \delta^k_i \right ] \nonumber
\end{align}
Note that the terms $\tilde \delta_i^k$ and $S_i^{k}$ do not depend on stage $k$ and remain constant when solving for $f_p^k$ and $f_n^k$. For the ease of notation, we define a new term $\mathbf C_r$ that indicates if $\bx_i$ is rejected at a $k$th stage. The term is close to one if $f^k_p$ and $f^k_n$ disagree (reject) and small if they agree.
 \[ \mathbf{C}_r(f^k_p,f^k_n,\bx^k_i,y_i)= \mathbf C(y_i f^k_p(\bx_i))+ \mathbf C(y_i f^k_n(\bx_i)) - 2 \mathbf C(y_i f^k_p(\bx_i)) \mathbf C(y_i f^k_n(\bx_i)) \]
The expressions for state variables and cost-to-go are now simplified. 
\begin{align}
S_i^{k+1}=S_i^{k} \mathbf{C}_r(f^k_p,f^k_n,\bx^k_i,y),~S_i^1=1 \label{eq:Sik}
\end{align}
The state variable remains greater than zero as long as $\bx_i$ is rejected at every stage. The expression for cost-to-go at $k$th stage is:
\begin{align}
\tilde \delta^{k}_i =  \underbrace{\delta^{k+1}}_{\mbox{\small meas. cost}}+\underbrace{\mathbf C(y_if^{k+1}_p(\bx^{k+1}_i))\mathbf C(y_if^{k+1}_n(\bx^{k+1}_i))}_{\mbox{\small err. penalty if not rejected at stage $k+1$}} +\underbrace{\tilde \delta^{k+1}_i \mathbf{C}_r(f^{k+1}_p,f^{k+1}_n,\bx^{k+1}_i,y)}_{\mbox{\small cost-to-to if rejected at stage $k+1$}} \label{eq:deltaik}
\end{align}
The last two terms are simply a surrogate for $L_k(\cdot)$ from \ref{eq:Lk} in terms of $\mathbf C(\cdot)$.

For the last stage (a standard binary classifier), we fix the first $K-1$ stages and solve:
\begin{eqnarray}
f^K=\arg \min_{q_1, q_2, \ldots} \sum_{i=1}^N S_i^K \mathbf C \left (y_i \sum_{h_j \in \cH^K} q_j h_j(\bx_i)\right ) \label{eq:fK}
\end{eqnarray}

Our algorithms performs cyclical optimization over the stages. To initialize $f^k_n,~f^k_p~\forall k$, we simply hard code $f^k_p$ to classify any $\bx$ as +1 and $f^k_n$ as -1 so that all $\bx$'s are rejected to the last stage. Using these nominal classifiers, we compute $S^k_i$ and $\delta^k_i$ according to equations \ref{eq:Sik} and \ref{eq:deltaik}, respectively. 

At a stage $k$, for a fixed $\delta^k_i$ and $S^k_i$, we alternate among minimizing $f^k_p$ and $f^k_n$ according to equations $\ref{eq:fp}$ and $\ref{eq:fn}$. In practice, we found that one iteration is sufficient. 

Given a new estimate of stage $k$, we update $\delta^s_i$ for $s>k$ and $S^s_j$ for $s<k$ and then move on to optimizing another stage $k'$. Given an estimate for stage $k'$, we again update the state variables and cost-to-go for the rest of the system. 

The stages are optimized in the following order. We start with the last stage and make our way backwards to the first stage. Then do a forward pass from $1$st stage to last. These forward and back passes are repeated it until convergence. See Algorithm \ref{alg:multistage_global}.

\begin{algorithm}[htb!]
\caption{Global Algorithm}
\label{alg:multistage_global}
\begin{algorithmic}
\STATE INPUT: $\{x_i,y_i\}_{i=1}^N$, $\{ \cH_k \}_{k=1}^K$ \COMMENT{Weak Learners for each stage}, $\{ \delta_k \}_{k=1}^{K}$ \COMMENT{costs}, $D$ \COMMENT{ Loop Iterations}
\STATE INITIALIZE: $f^k_n(x) \leftarrow +1,f^k_p(x) \leftarrow -1,~for~k=1 \ldots K-1$  \COMMENT{first $K-1$ stages reject everything}
\FOR {$d=1, \ldots,  D$}
\FOR{$k=K, \ldots, 1, 2, \dots K-1 $}
\STATE \COMMENT{Start from the last stage then iterate to the first stage and then back to last stage}
\IF{$k<K$}
\STATE FInd $f_p^k$ by solving boosting subproblem in \ref{eq:fp}
\STATE Find $f_n^k$ by solving boosting subproblem to \ref{eq:fn}
\ELSIF{$k=K$}
\STATE \COMMENT{Last Stage}
\STATE Find $f^K(x)$ by solving boosting subproblem in \ref{eq:fK} 
\ENDIF
\STATE Update $\tilde \delta^s_i$ for $s>k$ and $S^s_i$ for $s<k$
\ENDFOR
\ENDFOR
\STATE $F^k(\bx^k) \leftarrow \begin{cases} \sgn{f^k_p(\bx^k)}, & \mbox{if } \sgn{f^k_p(\bx^k)}=\sgn{f^k_n(\bx^k)} \\ \mbox{reject},& \mbox{if } \sgn{f^k_p(\bx^k)} \neq \sgn{f^k_n(\bx)}  \end{cases}$
\STATE OUTPUT: $F^1,F^2, \ldots, F^K$
\end{algorithmic}
\end{algorithm}

Our formulation allows us to form a surrogate for the entire risk in Equation \ref{eq:single}, not just for each subproblem. This enables us to prove the following theorem,
\newtheorem{thm3}[thm1]{Theorem}
\begin{thm3}
Our global surrogate algorithm converges to a local minimum.
\end{thm3}

\begin{proof} 
This is simply due to a fact that we are minimizing a global smooth cost function by coordinate descent over $\bq_p^1, \bq^1_n, \bq_p^2, \bq^2_n, \ldots, \bq^K$. Here, $\bq^k_p$ is the vector of weak learner weights parametrizing $f^k_p$. For the derivation of three stage system global cost refer to Appendix 8.
\end{proof}
However, since the global loss and the loss for each subproblem are non-convex programs, there is no global optimality guarantee. Theorem 3 ensures that our algorithm terminates. 

\paragraph{Regularization to reduce overfitting:}
To reduce overtraining, we introduce a simple but effective regularization. For any loss $\mathbf C(z)$ and a parameter $\lambda$, we introduce a multiplicative term to the cost function:$\min_{\bq} exp(\lambda |\bq|) \sum_{i=1}^N \mathbf C(y_i \sum_{h_j \in \cH} q_j h_j(\bx_i))$ The term $\exp(\lambda |\bq|)$ limits how large a step size for a weak hypothesis can become. It also introduces a simple stopping criteria: abort if ${\sum_{i=1}^n C'(y_i f_t(x_i)) y_i h_{t+1}(x_i) \over \sum_{i=1}^n \mathbf C(y_i f_t(x_i))}  \leq \lambda $. This corresponds to a situation when no descent directions ( weak hypothesis $h_{t+1}$ ) can be found to minimize the cost function.

\section{Generalization Error}
Our system is composed of margin maximizing classifiers, therefore it is appropriate to derive generalization error bounds based on margins. It turns out that we can employ maximum margin generalization techniques from \cite{bartlett98} to derive error bounds for a two stage version of the system. A two stage system consists of three boosted binary classifiers: 
\begin{align*}
f^1_p(\bx^1)=\sum_{h_j \in \cH^1} q^p_j h_j(\bx^1),~~f^1_n(\bx^1)=\sum_{h_j \in \cH^1} q^n_j h_j(\bx^1),~~f^2(\bx^2)=\sum_{h_j \in \cH^2} q^2_j h_j(\bx^2)  
\end{align*}
\newtheorem{thm4}[thm1]{Theorem}
\begin{thm4}
Let $\cD$ be a distribution on $\cX \times \{+1,-1\}$, and let $\cS$ be a sample of $m$ examples chosen independently at random according  to $\cD$, and a rejected subsample of size $m_r$, $\cS_r=\{x \in \cS| f^1_p(\bx) \neq f^1_n(\bx) \}$ Assume that the base-classifier spaces $\cH_1$ and $\cH_2$ are finite, and let $\delta>0$. Then with probability at least $1 - \delta$ over the random choice of the training set $S$, all boosted classifiers $f^1_n,f^1_p,f^2$ satisfy the following bound for all $\theta_1 > 0$ and $\theta_2 >0$:
\begin{align}
&\Pr_{\cD}[yf^1_n(\bx) \leq 0, yf^1_p(\bx) \leq 0] \nonumber  + \Pr_{\cD} [ yf^2(\bx) \leq 0, f^1_n(\bx) \neq f^1_p(\bx) ] \leq \nonumber \\
&\Pr_{\cS}[yf^1_n(\bx) \leq \theta_1, yf^1_p(\bx) \leq \theta_1] + \Pr_{\cS_r}[yf^2(\bx) \leq \theta_2] + \nonumber \\
&\mathcal O \left ( {1\over \sqrt m} \left ( { \log m \log |\cH_1| \over \theta_1} + \log {1 \over \delta} \right)^{1\over2} \right )
+ \mathcal O \left ( {1\over \sqrt m_r} \left ( { \log m_r \log |\cH_2| \over \theta_2} + \log {1 \over \delta} \right)^{1\over2} \right )
\end{align}
\end{thm4}
\begin{proof}
The proof extends the approach in \cite{bartlett98} to a two stage system. For complete details please refers to the appendix.
\end{proof}
The two stage system can be compactly expressed:
\begin{align}
F(\bx)=
\begin{cases}
\sgn{f^1_p(\bx^1)},& \sgn{f^1_p(\bx^1)}=\sgn{f^1_n(\bx^1)} \\
\sgn{f^2(\bx^2)},& \sgn{f^1_p(\bx^1)} \neq \sgn{f^1_n(\bx^1)}
\end{cases}
\end{align}
The system error is a sum of two terms: error at the 1st stage + error at the 2nd stage. Theorem 4 states the generalization error of $F(\bx)$ is bounded by the empirical margin error over the training set $S$ plus a term that is inversely proportinal to the margins and the number of training samples at that stage.
An interesting observation is that $m_r$, number of samples that reaches the 2nd stage, depends on the reject classifier at the 1st stage. So if very few examples make it to the second stage then we do not have strong generalization.

\section{Experiments}
The goal is to demonstrate that a large fraction of data can be classified at an early stage using a cheap modality. In our experiments, we use four real life datasets with measurements arising from meaningful stages. 
\subsection{Related Algorithms:} We compare our algorithm to two methods:
\paragraph{Myopic:} An absolute margin of a classifier is a measure of how confident a classifier is on an example. Examples with small margin have low confidence and should be rejected to the next stage to acquire more features. This approach is based on reject classification (\cite{barlett08}). We know from Claim 1 that the optimal classifier is a threshold of the posterior. For each stage, we obtain a binary boosted classifier, $f^k(\cdot)$, trained on all the data. We then threshold the margin of the classifier, $|f^k(\bx^k)|$. It is known that given an infinite amount of training data, boosting certain losses (sigmoid loss in our case) approaches the log likelihood ratio, $f(\bx)={1 \over 2} \log{ \Pr(y=1|\bx) \over \Pr(y=-1|\bx)}$(\cite{vasc09}). So a reject region for a given threshold $t_k$ is defined: $\{\bx \mid |f^k(\bx)| \leq t_k\}$. This is a completely myopic approach as the rejection does not take into account performance of later stages. This method is very similar to TEFE (\cite{liu08}) which also uses absolute margin as a measure for rejection. The difference is that our myopic strategy is a boosting classifier not an SVM as used in TEFE.

\paragraph{Expected Utility/Margin:}  An expected margin difference measures how a new attribute, if acquired, would be useful for an example. If this expected utility for an example is large then a new attribute should be acquired. This approach is based on the work by \cite{kanani:2008}. We train boosted binary classifiers on all the data for each stage: $f^k(\bx^k)$. Given the measurement at the current stage $\bx^k$, we compute an expected utility (change in normalized margin) of acquiring the next measurement $\bx_{k+1}$:
\[
U(\bx^k)=\sum_{x_{k+1} \in \mathcal X_{k+1}} \left |  f^{k}(\bx^k)- f^{k+1}([\bx^k,x_{k+1}])  \right |  \Pr(x_{k+1} | \bx_k)
\]
An $\bx^k$ is rejected to the next stage if its utility $U(\bx^k) \geq t_k$ is greater than a threshold. Here, $\mathcal X_{k+1}$ denotes the possible values that $x_{k+1}$ can take. Note this approach requires estimating $\Pr(x_{k+1}|\bx^k)$\footnote{While there are many different ways to estimate a probability likelihood we used a Gaussian mixture due to its computational efficiency}, therefore the $(k+1)$th measurement has to be discrete or distribution needs to be parametrized.  Due to this limitation, we only compare this method on two datasets. 

\subsection{Simulations}
\paragraph{Performance Metric: } A natural performance metric is the trade off between system error and measurement cost. Note, for utility and myopic methods, it is unclear how to set a thresholds $t_k$ for each stage given a measurement cost $\delta_k$. For this reason, we only compare them in a two stages system. More than two stages is not-practical because we would need to test every possible $t_k$ for every stage $k$. In a two stage setting, measurement cost is proportional to the fraction of examples rejected to the second stage. For our algorithm, we vary a reject cost $\delta$ to generate a system error vs reject rate plot. For margin and utility, we sweep a threshold $t_k$. System error is the sum of $1$st stage and $2$nd stage errors. Reject rate is the fraction of examples rejected to the 2nd stage and require additional measurements. Low reject rate (cost) corresponds to higher error rate as most of the data will be classified at the first stage using less informative measurements. High reject rate will have performance similar to a centralized classifier, as most examples will be classified at the 2nd stage.

\paragraph{Set Up:}In all our experiments, we use stumps \footnote{stump classifier is threshold on $d$th dimension: $h_{d,g,\{+1/-1\}}(x)=\{+1/-1\} sign (x(d)-g)$} as weak learners. For each dataset and experiment, we randomly split the data $50/50$ for training and testing. The results are evaluated on a separate test set, and the simulations are averaged over $50$ monte-carlo trials. The number of iterations for each boosting subproblem is set to $T=50$. In our global surrogate algorithm, the number of outer loop iterations is set to $D=10$ 

\begin{table}[htb!]
\centering
\begin{tabular}{| l | c | c | c | c }
\hline
Name & Size & $1$st Stage & $2$nd Stage \\ \hline \hline
Gassian Mixture & $1000$ & $1$st dim & $2$nd dim \\ \hline
Mammogram Mass & $830$  & $3$ CAD meas. & Radioligist Rating \\ \hline
Pima Diabetes  & $810$ & $6$ simple tests: BMI, sex, ..& $2$ blood tests  \\ \hline
Polyps & 310 & $12$ freq. bins &  $126$ freq. bins \\ \hline 
Threat & 1300 & Images in IR, PMMW & Images in AMMW \\ \hline
\end{tabular}
\caption{Dataset Descriptions}
\end{table}

\paragraph{Discrete Valued Data Experiments:} To compare our method to the utility approach, we consider discrete data. The first dataset is a quantized (with 20 levels) Gaussian mixture synthetic data in two dimension. The 1st dimension is stage one; the 2nd dimension is stage two. The second dataset is Mammogram Mass from UCI Machine Learning Repository. It is used to predict the severity of a mammographic mass lesion (malicious or benign). It contains $3$ attributes extracted from the CAD image and also an evaluation by a radiologist on a confidence scale in addition to the true biopsy results. 
The first stage are features extracted from the CAD image, and the second stage is the expert confidence rated on a discrete scale $1-5$. Automatic analysis of the CAD image is cheaper than employing an opinion of a radiologist.

\begin{figure}[htb!]
\begin{center}
\subfigure[GaussMixQuant]{\includegraphics[width=.32 \linewidth]{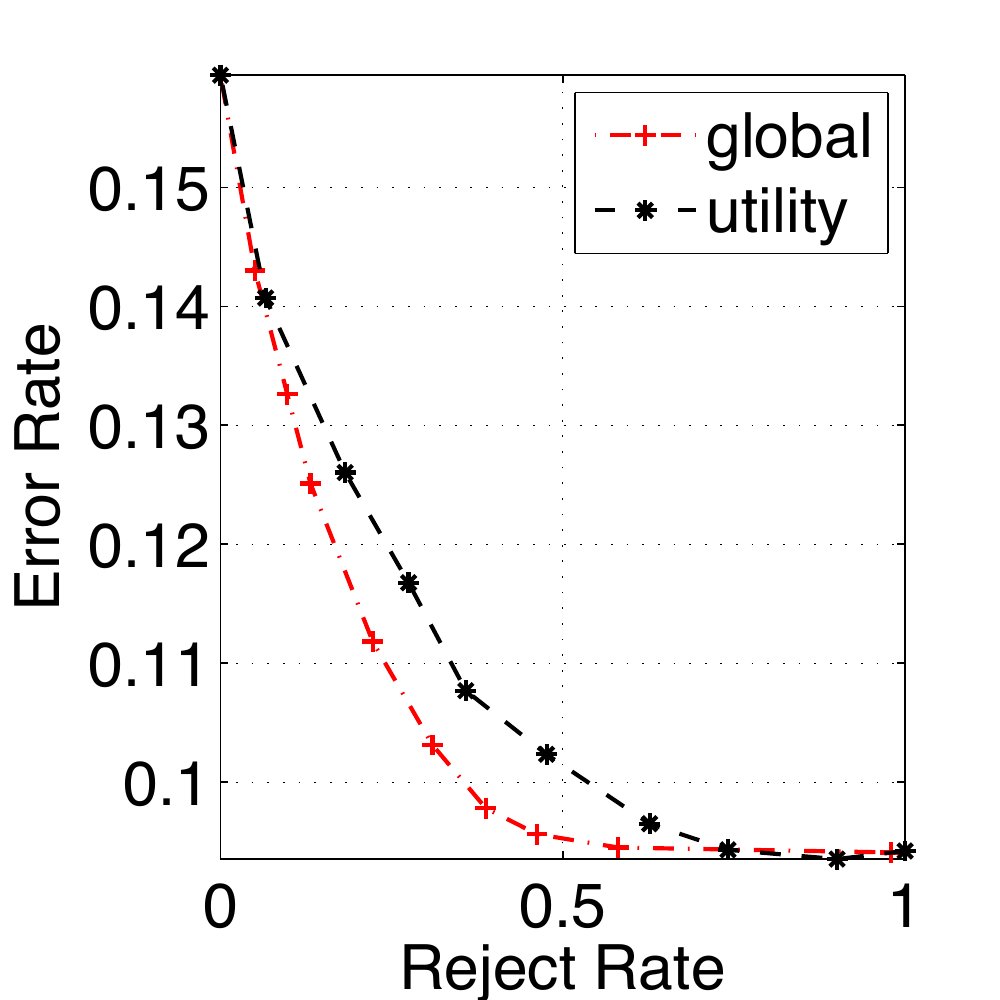}}
\subfigure[mammogram]{\includegraphics[width=.32 \linewidth]{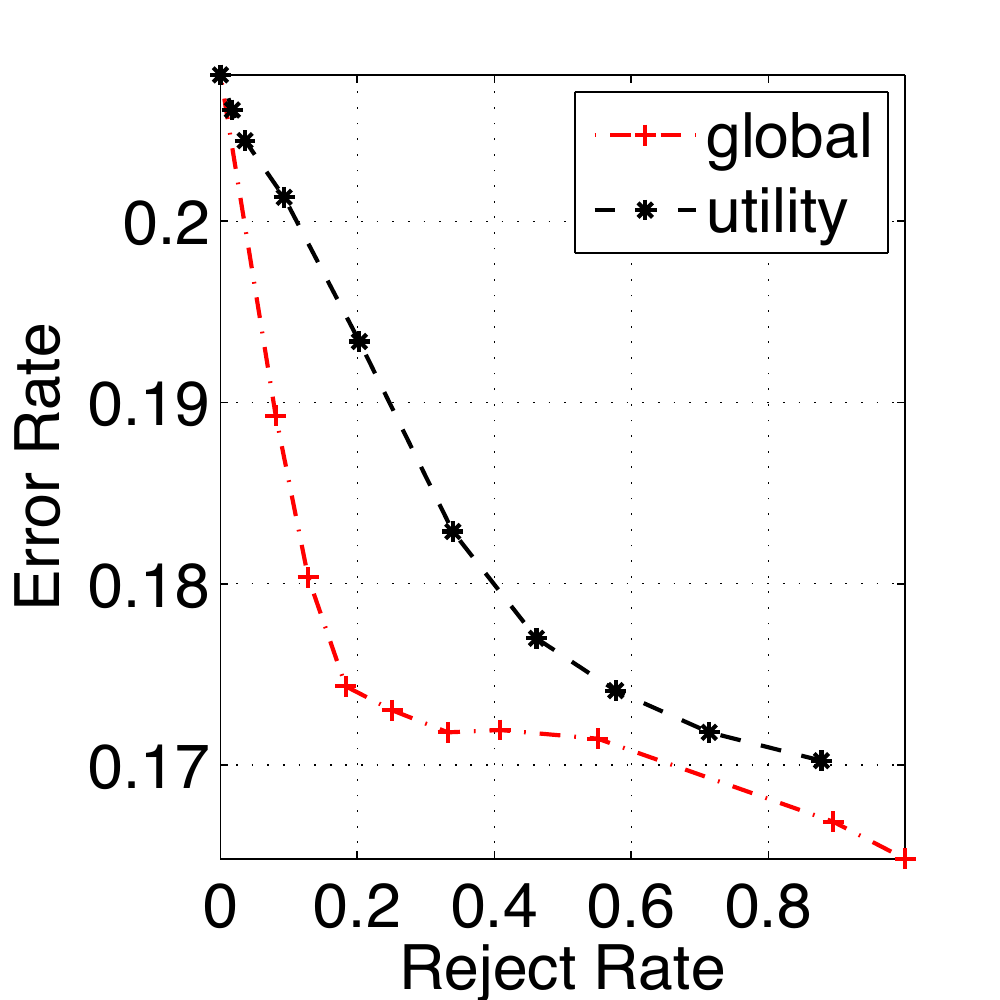}}
\caption{Comparison of Global to Utility on (a) quantized two gaussian clusters and (b) mammogram dataset. Reject Rate vs System Error. Reject Rate is the fraction of examples with measurements from both stages. Our approach outperforms Utility possibly because we do not need to estimate probability likelihoods}
\label{fig:utility}
\end{center}
\end{figure}

Simulations in Fig. \ref{fig:utility} demonstrate that utility performs worse when compared to our approach. This is possibly due to poor probability estimates in limited data setting.

\paragraph{Continuous Valued Data Experiments} We compare our global method to the myopic method on three datasets. The Pima Indians Diabetes Dataset (UCI MLR) consists of $8$ measurements. $6$ of the measurements are inexpensive to acquire and consist of simple tests such as body mass index, age, pedigree. These we designate as the first stage. The other two measurements constitute the second stage and require more expensive procedures.

The polyp dataset consists of hyper-spectral measurements of colon polyps collected during colonoscopies (\cite{diaz09}). The attribute is a measured intensity at 126 equally spaced frequencies. Finer resolution requires higher photon count which is proportional to acquisition time. For a first stage, we use a coarse measurement downsampled to only 12 frequency bins. The second stage is the full resolution frequency response. Using the course measurements is cheaper than acquiring the full resolution. 

The threat dataset contains images taken of people wearing various explosives devices. The imaging is done in three modalities: infrared (IR), passive millimeter wave (PMMW), and active millimeter (AMMW). All the images are registered. We extract many patches from the images and use them as our training data.  A patch carries a binary label, it either contains a threat or is clean. IR and PMMW are the fastest modalities but also less informative. AMMW requires raster scanning a person and is slow but also the most useful.
\def \fw {.32}
\begin{figure}[htb!]
\begin{center}
\subfigure[pima]{\includegraphics[width=\fw \linewidth]{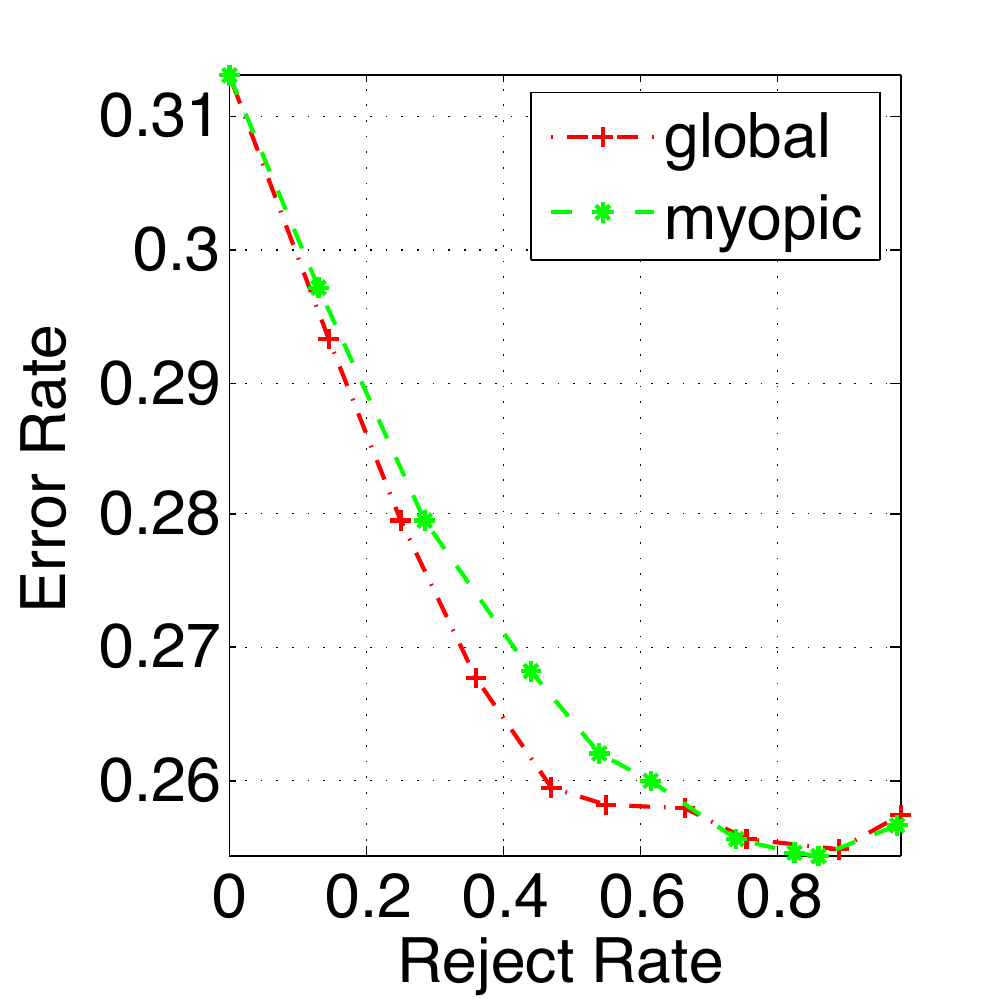}}
\subfigure[polyps]{\includegraphics[width=\fw \linewidth]{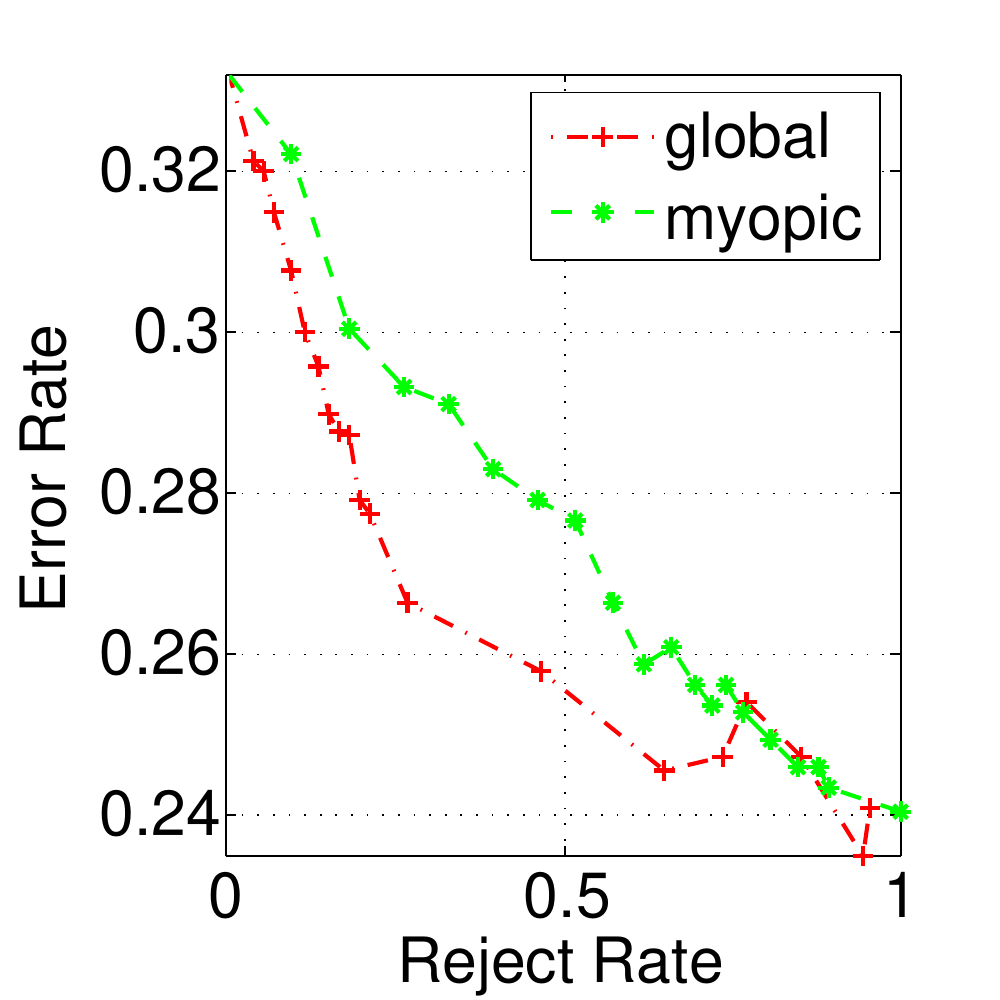}}
\subfigure[threat]{\includegraphics[width=\fw \linewidth]{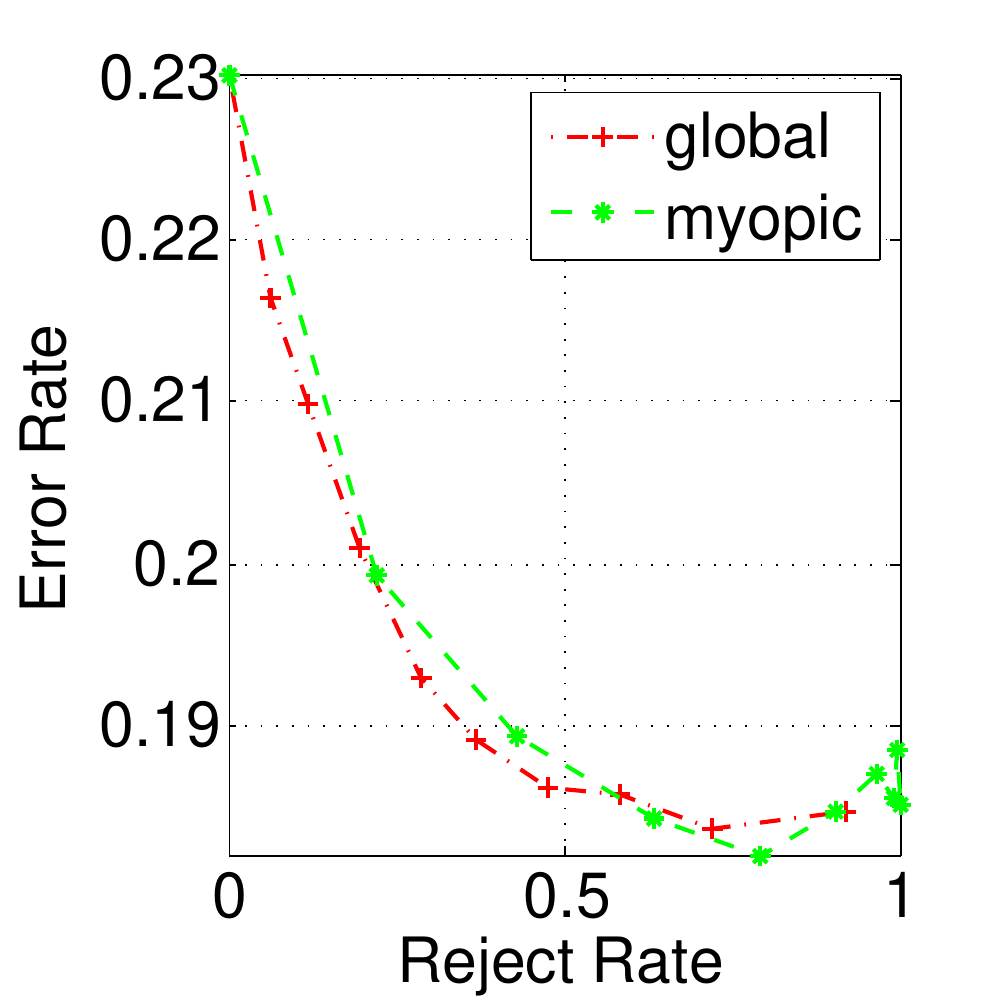}}
\caption{ Three datasets are evaluated: pima, polyps and threat. Reject Rate vs Error Rate for a varying reject cost $\delta$. Reject Rate is the fraction of examples with measurements from both stages. Global and Myopic are compared. Global (our approach) has a better performance over all while Myopic does better in some situations.}
\label{fig:cont}
\end{center}
\end{figure}

\begin{table}[htb!]
\centering
\begin{tabular}{| c | c | c | c | c |}
\hline
Name & Centralized & Utility & Myopic  & Ours \\ \hline \hline
2D Gaussian Mix & 0.09&50\%&- &30\%\\ \hline
Mammogram &0.165&60\%&-&15\% \\ \hline
Pima Diabetes  & 0.26 &-&60\%&45\%\\ \hline
Polyps &0.24&-&75\%&50\%\\ \hline 
Threat  &0.185&-&50\%&45\%\\ \hline
\end{tabular}
\caption{Performance illustration for different datasets (quantitate view of the curves). Datasets have 2 sensing modalities. Centralized denotes the test error obtained with all modalities. Last three columns denotes performance for different approaches. Performance is measured by the average number of examples requiring 2nd stage to achieve error close to centralized. Utility approach does not work for last three datasets due to high-dimensionality issues. We note the significant gains of our approach over competing ones of many interesting datasets.}
\end{table}

In Fig. \ref{fig:cont}, global performs better than margin in most cases. On threat data, margin appears to be doing just marginally worse than global, however, we get only a few points on the curve with reject rates less than $50\%$. Due to the heuristic nature of margin, we cannot construct a multistage classifier with an arbitrary reject rate. 

The goal is to reach the performance of a centralized classifier ($100 \%$ reject rate) while utilizing the 2nd stage sensor only for a small fraction of examples. Overall, the results demonstrate the benefit of multi-stage classification: rejection rate can be set to less than~$50\%$ with only small sacrifices in performance. For the mammogram data, this implies that for half of the patients a diagnoses can be made solely by an automatic analysis of a CAD image without an expensive opinion of a radiologist. For the Pima data, similar error can be achieved without an expensive medical procedures. For the polyps dataset, a fast low resolution measurement is enough to classify a large fraction of patience. In the threat dataset, IR and PMMW are sufficient to decide whether or not a threat is present for the majority of instances without requiring a person to go through a slower AMMW scanner.

\paragraph{Unbalanced False Positive and False Negative Penalties:}
In medical diagnosis and threat detection, the penalty of false positives and false negatives is not equal. We can easily adapt our algorithm to account for such setting. Empirical Risk in \ref{eq:emprisk} can be modified to include a penalty of $w_p$ for a Type I error and $w_n$ for a Type II error. The experiment in Fig. \ref{fig:biased} demonstrates our global algorithms in such scenario. For each reject cost $\delta$, we compute an ROC curve. We also compute a corresponding average reject rate for each value of $\delta$. So the highest reject rate corresponds to the best performance but also to the highest acquisition cost incurred by the system. Note that very good performance can be achieved by requesting only 50\% of instances to be measured at the second stage.

\def \fw {.32}
\begin{figure}[htb!]
\begin{center}
\subfigure[pima]{\includegraphics[width=\fw \linewidth]{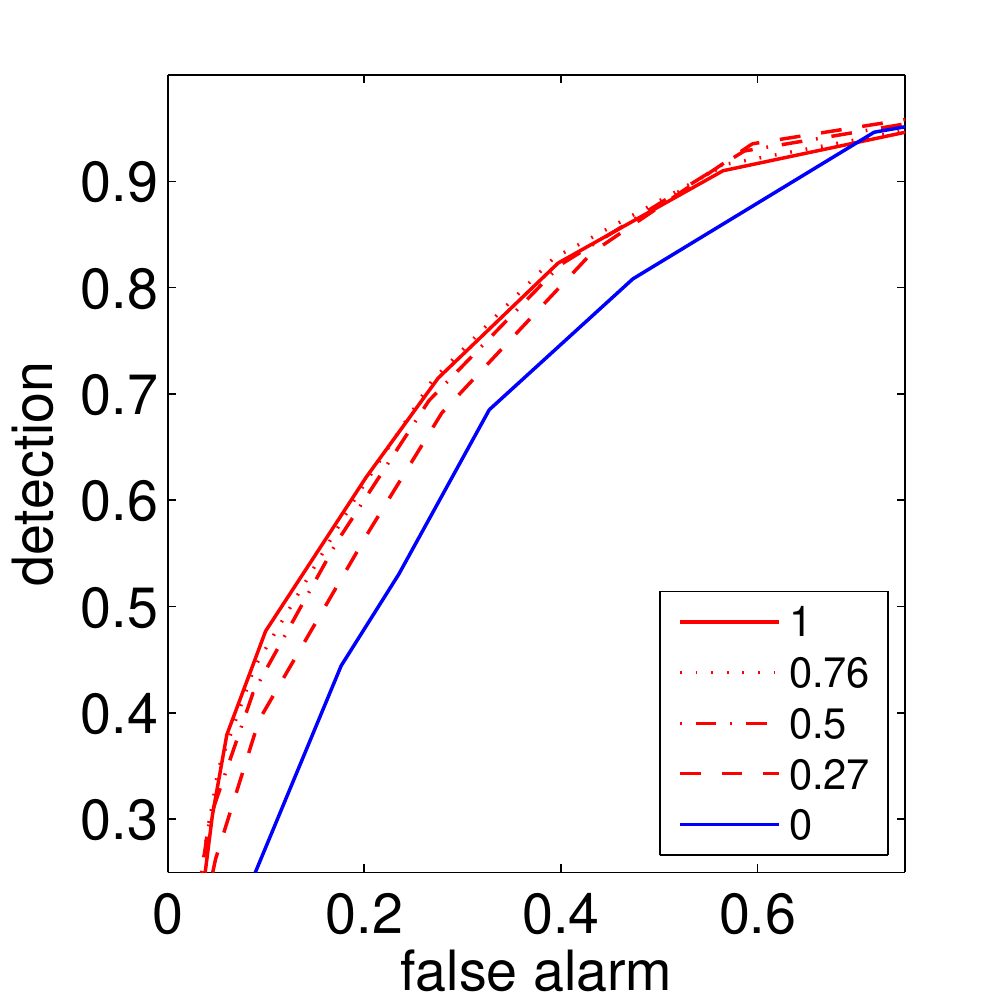}}
\subfigure[polyps]{\includegraphics[width=\fw \linewidth]{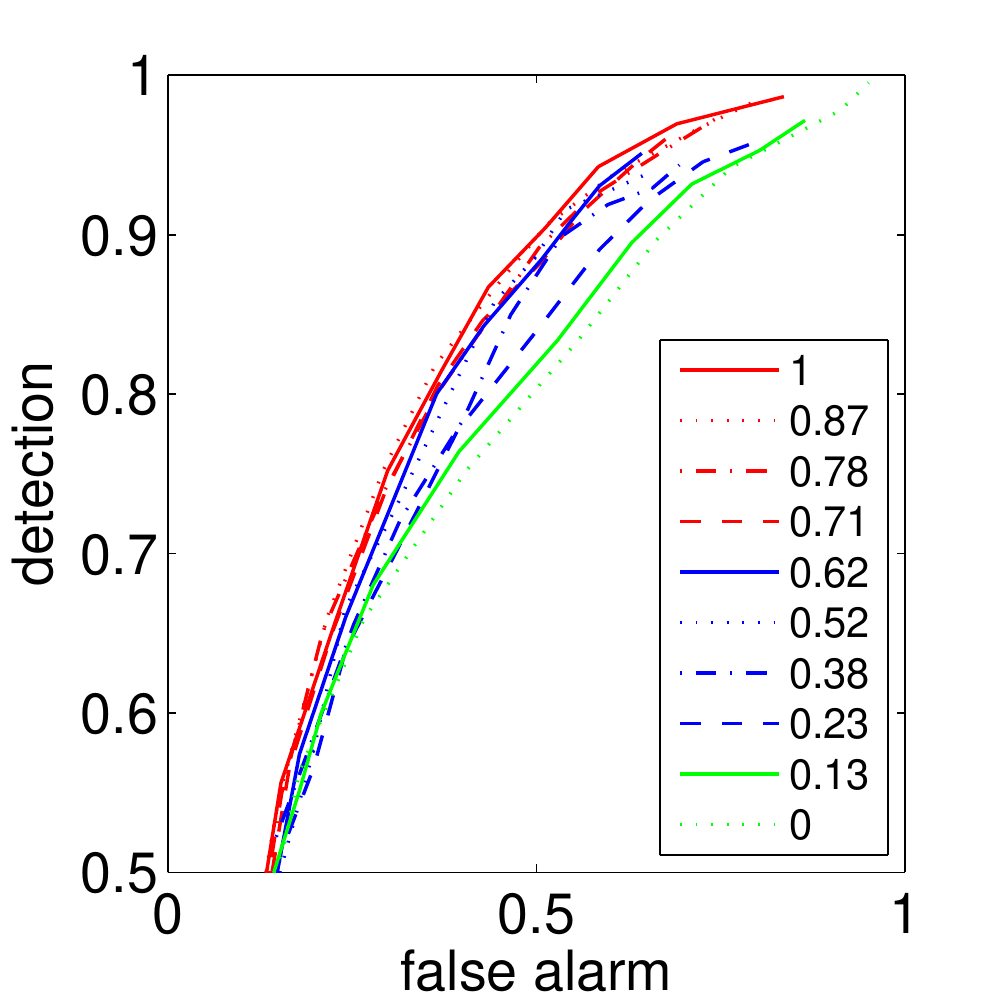}}
\subfigure[threat]{\includegraphics[width=\fw \linewidth]{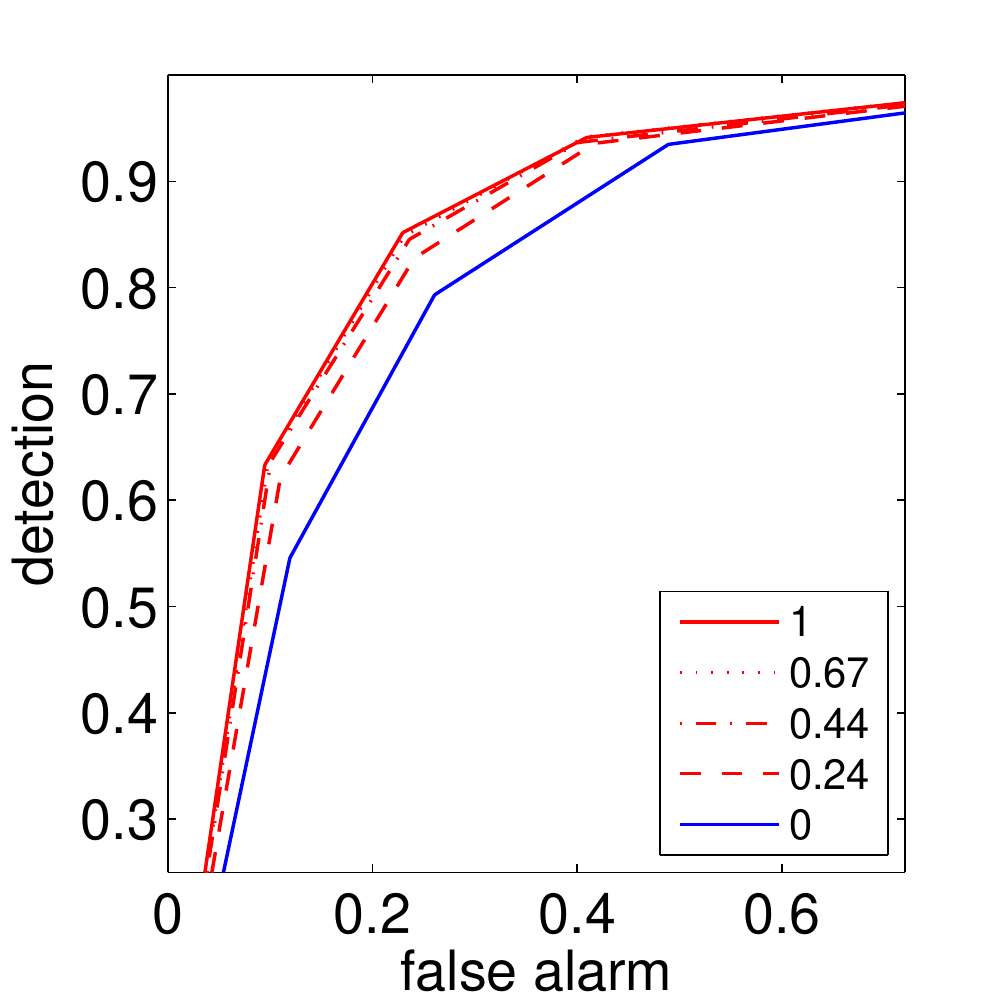}}
\caption{Two Stage ROC using the global surrogate method. Each ROC curve corresponds to a different value of reject cost $\delta$. The legend displays average reject rate for $\delta$'s. Note, the red ROC corresponds to the centralized system ($100 \%$ reject rate). Very good performance can be achieved by requesting only ~$50\%$ of instances to be measured at the second stage.}
\label{fig:biased}
\end{center}
\end{figure}

\begin{figure}[htb!]
\begin{center}
\includegraphics[width=.5 \linewidth]{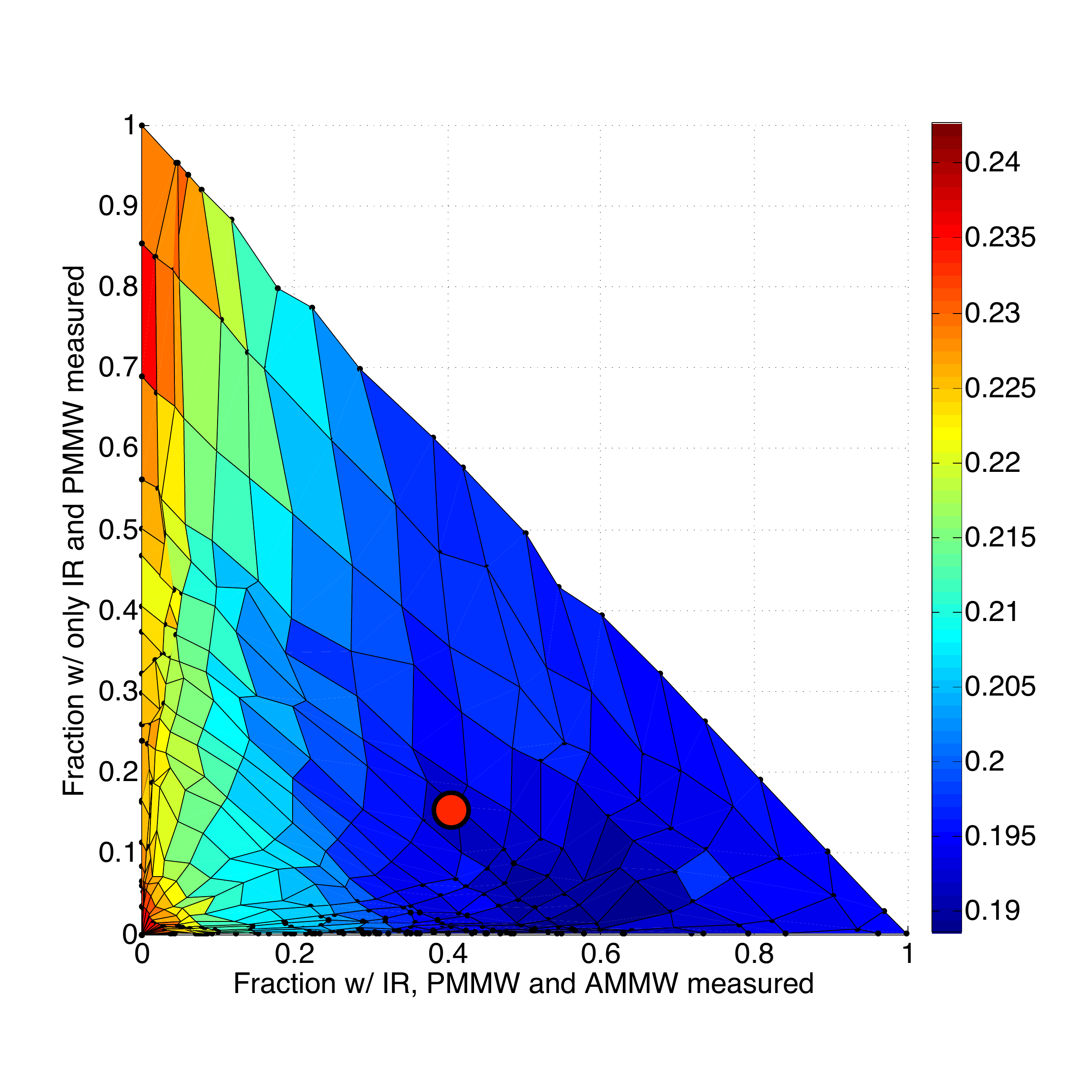}
\caption{Three Stage System. The color maps error. A point on the map corresponds to a performance of a particular multistage classification strategy. The vertical axis is the fraction of examples for which only IR and PMMW measurements are used in making a decision. The horizontal axis is the fraction of examples for which all three modalities are used. An example red point in the figure, $\{.4,.15,.195\}$, correspond to a system where $40\%$ of examples use IR and PMMW, $15\%$ use only $IR$ and the rest of data ($45\%$) use all the modalities. And this strategy achieves a system error rate of $19.5\%$. }
\label{fig:gauss3}
\end{center}
\end{figure}

\paragraph{Three Stages:} Lastly, we demonstrate a three stage system, we apply our algorithm to three stages of threat dataset. Note for margin it is unclear how to generalize it to a multistage scenario and there is no way to define reject costs for different stages. We set the first stage to be IR, second PMMW and AMMW as third. There is no cost for acquiring $IR$. We vary the costs for the PMMW (2nd) stage, $\delta_1$, and AMMW (3rd), $\delta_2$, to generate an error map (color in Fig. \ref{fig:gauss3}). A point on the map corresponds to a performance of a particular multistage classification strategy. The vertical axis is the fraction of examples for which only IR and PMMW measurements are used in making a decision. The horizontal axis is the fraction of examples for which all three modalities are used. For example, a red point in the figure, $\{.4,.15,.195\}$, correspond to a system where $40\%$ of examples use IR and PMMW, $15\%$ use only $IR$ and the rest of data ($45\%$) use all the modalities. And this strategy achieves a system error rate of $19.5\%$. Note that the support lies below the diagonal. This is because the sum or reject rates has to be less than one. Results demonstrate some interesting observations. While best performance (about $19\%$) is achieved when all the modalities are used for every example, we can move along the vertical lines and allow a fraction to be classified by IR and PMMW, avoiding AMMW all together. This strategy achieves performance comparable to a centralized system, (IR+PMMW+AMMW).

\section{Conclusion}
In this paper, we propose a general framework for a sequential decision system in a non-parametric setting. Starting from basic principles, we derive the bayesian optimal solution. Then, to simplify the problem, we parameterize a classifier at each stage in terms of two binary decisions. We formulate an ERM problem and optimize it by alternatively minimizing one stage at a time. Remarkably, all subproblems turn out to be weighed binary error minimizations. We introduce a practical boosting algorithm that minimizes a global surrogate of the empirical risk and test it on several datasets. Results show the advantage of our formulation to more heuristic approaches. Overall, our experiments demonstrate how multi-stage classifiers can achieve good performance by acquiring full measurements only for a fraction of samples.

\begin{acknowledgements}
This work is supported by the U.S. Department of Homeland Security 
2008-ST-061-ED0001
\end{acknowledgements}

\nocite{trapeznikov:2012}

\bibliographystyle{apalike} 

\bibliography{bibliography}

\section{Appendix}
\subsection{Proof of Theorem 4}
\begin{proof}
This will closely follow the proof of Theorem 1 in  \cite{bartlett98}.
We have to bound two terms:
\[\Pr_{\cD}[yf_n(x) \leq \theta_1,yf_p(x) \leq \theta_1] \mbox{ and } \Pr_{\cD} [ yf_2(x) \leq \theta_2, yf_n(x) \neq yf_p(x)]\]

{\bf First Term} Let us bound the first term. Define $\cC_N$ to be the set of unweighted averages over $N$ elements from $\cH_1$,
\begin{align}
\cC_N=\{f: x \rightarrow {1 \over N} \sum_{i=1}^N h_i(x) \mid  h_i \in \cH_1 \}
\end{align}
Any weighed classifier $f=\sum_{h} q_h h(x)$ can be approximated by drawing an element from $\cC_N$ by choosing $h_1... h_N$ with prob. $q_h$.

We can express our first term as a sum of probabilities of disjoint events.
\begin{align}
\Pr_{\cD} \left [yf_p(x) \leq 0,yf_n(x) \leq 0 \right ] = \\
\Pr_{\cD} \left [yf_p(x) \leq 0,yf_n(x) \leq 0, yg_p(x) \leq { \theta_1 \over 2}, yg_n(x) \leq { \theta_1 \over 2}  \right ] \\
+ \Pr_{\cD} \left [yf_p(x) \leq 0,yf_n(x) \leq 0, yg_p(x) \leq { \theta_1 \over 2}, yg_n(x) > { \theta_1 \over 2}  \right ]  \\
+ \Pr_{\cD} \left [yf_p(x) \leq 0,yf_n(x) \leq 0, yg_p(x)  > { \theta_1 \over 2}, yg_n(x) \leq { \theta_1 \over 2}  \right ] \\
+ \Pr_{\cD} \left [yf_p(x) \leq 0,yf_n(x) \leq 0, yg_p(x) >  { \theta_1 \over 2},  yg_n(x) >  { \theta_1 \over 2} \right ]
\end{align}
Further, we can write,
\begin{align}
\Pr_{\cD} \left [yf_p(x) \leq 0,yf_n(x) \leq 0 \right ] \leq
\Pr_{\cD} \left [ yg_p(x) \leq { \theta_1 \over 2}, yg_n(x) \leq { \theta_1 \over 2} \right ]  \\+ \Pr_{\cD} \left [yf_p(x) \leq 0,yf_n(x) \leq 0, yg_p(x) >  { \theta_1 \over 2},  yg_n(x) >  { \theta_1 \over 2} \right ] \label{eq5}
\end{align}
The inequality holds for any $g_p,g_n$. We take the expected value of the right hand side wrt to the distribution $\cC$
\begin{align}
\Pr_{\cD} \left [yf_p(x) \leq 0,yf_n(x) \leq 0 \right ] \leq \\
\E_{\cC} \left [\Pr_{\cD} \left [ yg_p(x) \leq { \theta_1 \over 2}, yg_n(x) \leq { \theta_1 \over 2} \right ] \right ]\\
+\ev{\cD}{\prob{\cC_p,\cC_n}{yg_p(x) >  { \theta_1 \over 2},  yg_n(x) >  { \theta_1 \over 2} \mid yf_p(x) \leq 0,yf_n(x) \leq 0 }}
\end{align}
The last term inside the expectation is the probability that an average of $N$ bernoulli random variables is larger than its expectation, we use a concentration result from Equation (4) in Theorem 1 of \cite{bartlett98}.
\begin{align}
\prob{\cC_p,\cC_n}{yg_p(x) >  { \theta_1 \over 2},  yg_n(x) >  { \theta_1 \over 2} \mid yf_p(x) \leq 0,yf_n(x) \leq 0 } \leq exp \left ({-N\theta_1^2 \over 8}\right ) \label{eq:bern}
\end{align}
To bound the first we use the result from Equation (5) in Theorem 1 of \cite{bartlett98}. if we set $\epsilon_N=\sqrt{(1/2m) \log ((N+1) |\cH_1|^{2N}) /\delta_N}$, with probability at least $1-\delta_N$,
\begin{align}
\prob{\cD,\cC}{yg_p(x) \leq { \theta_1 \over 2},yg_n(x) \leq { \theta_1 \over 2}} \leq \prob{S,\cC}{yg_p(x) \leq { \theta_1 \over 2},yg_n(x) \leq { \theta_1 \over 2}} + \epsilon_N
\end{align}
for any choice of $\theta$ and every distribution $\cC$. Here, $\prob{S}{}$ is probability taken with respect to a randomly drawn sample of size $m$ from $\cD$.

By the same argument as in inequality \ref{eq5},
\begin{align}
\prob{S,\cC_p}{yg_p(x) \leq { \theta_1 \over 2},yg_n(x) \leq { \theta_1 \over 2}} \leq \\
\prob{S}{yf_p(x) \leq \theta_1,yf_n(x) \leq \theta_1} + \ev{S}{\prob{\cC_p}{yg_p(x) \leq { \theta_1 \over 2} \mid yf_p(x) > \theta}}
\end{align}
The expressions inside the expectation can be bounded using the same Chernoff bound result from \ref{eq:bern},
\begin{align}
\prob{\cC}{yg_p(x) \leq { \theta_1 \over 2}, yg_n(x) \leq { \theta_1 \over 2} \mid yf_p(x) > \theta_1, yf_p(x) > \theta_1} \leq exp \left ({-N\theta_1^2 \over 8}\right )
\end{align}
By setting $\delta_N=\delta / (N(N+1))$, and combining the terms,
\begin{align}
\prob{\cD}{yf_p(x) \leq 0, yf_n(x) \leq 0} \leq \\ \prob{S}{yf_p(x) \leq \theta_1, yf_n(x) \leq \theta_1}+2exp \left ({-N\theta_1^2 \over 8}\right )+2\sqrt{{1 \over 2m} \log \left ({ N(N+1)^2|\cH_1|^{2N} \over \delta} \right )}
\end{align}
By setting, $N=(4/\theta^2_1) \log (m/ \log|\cH_1|^2)$,
\begin{align}
\prob{\cD}{yf_p(x) \leq 0, yf_n(x) \leq 0} \leq  \prob{S}{yf_p(x) \leq \theta_1, yf_n(x) \leq \theta_1} + \mathcal O \left ( {1\over \sqrt m} \left ( { \log m \log |\cH|^2 \over \theta} + \log {1 \over \delta} \right)^{1\over2} \right )
\end{align}

{\bf Second Term} Here we will bound the second term, $\Pr_{\cD} [ yf_2(x) \leq \theta_2, yf_n(x) \neq yf_p(x)]$
Define a new distribution:
\begin{align}
D_r= \begin{cases} cD(x,y),& f_p(x) \neq f_n(x) \\ 0, & f_p(x)=f_n(x) \end{cases}
\end{align}
Rewrite:
\begin{align}
\Pr_{\cD} [ yf_2(x) \leq \theta_2, yf_n(x) \neq yf_p(x)] \leq \Pr_{\cD} [ y f_2(x) \leq \theta_2  \mid yf_n(x) \neq yf_p(x) ]\\
= \Pr_{\cD_r} [ y f_2(x) \leq \theta_2]
\end{align}
Note that  $\cS_r$ is an iid  sample from $\cD_r$. Using Theorem 1 in \cite{bartlett98},
\begin{align}
\Pr_{\cD_r} [ yf_2(x) \leq 0 ] \leq \nonumber
\Pr_{\cS_r}[yf_2(x) \leq \theta_2] + \mathcal O \left ( {1\over \sqrt m} \left ( { \log m \log |\cH_2| \over \theta_2} + \log {1 \over \delta} \right)^{1\over2} \right )
\end{align}
Collecting the two terms produces the desired result.
\end{proof}

\def \bC {\mathbf C}
\section{Derivation of a global risk for a three stage system}
Consider a three stage system.
Define some terms:
\begin{align}
\mbox{Error Indicator: } \indicator{f(\bx)\neq y} \rightarrow \bC(yf(\bx))={1 \over 1+\exp(yf(\bx))} \\
\mbox{Reject Indicator: }\indicator{f_p(\bx)\neq f_n(\bx)} \rightarrow \\ \bC_r(f_p,f_n,\bx,y)=\bC(yf_p(\bx))+\bC(yf_n(\bx))-2\bC(yf_p(\bx))\bC(y_nf(\bx))
\end{align}
Risk for three stages:
\begin{align}
&R(f^1_p,f^1_n,f^2_p,f^2_n,f^3,\bx,y)=S^1R^1+S^2R^2+S^3R^3\\
&S^1=1\\
&S^2(f^1_p,f^1_n,\bx,y)=\bC_r(f^1_p,f^1_n,\bx^1,y)\\ 
&S^3(f^1_p,f^1_n,f^2_p,f^2_n,\bx,y)=\bC_r(f^1_p,f^1_n,\bx^1,y)\bC_r(f^2_p,f^2_n,\bx^2,y)\\
&R^1(f^1_p,f^1_n,\bx,y)=\bC(yf^1_p(\bx^1))\bC(yf^1_n(\bx^1))+\delta^2 \bC_r(f^1_p,f^1_n,\bx^1,y)\\
&R^2(f^2_p,f^2_n,\bx,y)=\bC(yf^2_p(\bx^2))\bC(yf^2_n(\bx^2))+\delta^3 \bC_r(f^2_p,f^2_n,\bx^2,y)\\
&R^3(f^3_p,\bx,y)=\bC(yf^3(\bx^3))\\
\end{align}
Plug in all the terms:
\begin{align}
R(\cdot)=& \underbrace{\bC(yf^1_p(\bx^1))\bC(yf^1_n(\bx^1))  + \delta^2 \bC_r(f^1_p,f^1_n,\bx^1,y)}_{R^1} \\
+ & \underbrace{ \bC_r(f^1_p,f^1_n,\bx^1,y)}_{S^2} \underbrace{\left \{ \bC(yf^2_p(\bx^2))\bC(yf^2_n(\bx^2))+\delta^3 \bC_r(f^2_p,f^2_n,\bx^2,y) \right \}}_{R^2} \\
+ & \underbrace{\bC_r(f^1_p,f^1_n,\bx^1,y)\bC_r(f^2_p,f^2_n,\bx^2,y)}_{S^3} \underbrace{\bC(yf^3(\bx^3))}_{R^3}
\end{align}
Minimize over $f_p^1,f^1_n$ and keep $f_p^2,f_n^2,f^3$ constant. We can rearrange the terms to get:
\begin{align}
&\arg \min_{f^1_p,f^1_n} \sum_{i} R(f^1_p,f^1_n,f^2_p,f^2_n,f^3,\bx_i,y_i)=\\
&\arg \min_{f^1_p,f^1_n} \sum_{i} \bC(yf^1_p(\bx_i^1))\bC(yf^1_n(\bx_i^1))  + \tilde \delta^1_i \bC_r(f^1_p,f^1_n,\bx^1,y) \\
&\mbox{such that:} \\
&\tilde \delta^1_i=\delta^2+\left \{ \bC(yf^2_p(\bx^2))\bC(yf^2_n(\bx^2))+\delta^3 \bC_r(f^2_p,f^2_n,\bx^2,y) \right \} \\&+ \bC_r(f^2_p,f^2_n,\bx^2,y)\bC(yf^3(\bx^3))
\end{align}
Minimize over $f^2_p,f^2_n$ and keep $f_p^1,f^1_n,f^3$ constant:
\begin{align}
&\arg \min_{f^2_p,f^2_n} \sum_{i} R(f^1_p,f^1_n,f^2_p,f^2_n,f^3,\bx_i,y_i)= \\
&\arg \min_{f^2_p,f^2_n} \sum_{i}S^2_i \left \{ \bC(yf^2_p(\bx_i^2))\bC(yf^2_n(\bx_i^2))  + \tilde \delta^2_i \bC_r(f^2_p,f^2_n,\bx_i^2,y) \right \} \\
&\mbox{such that:} \\
&S^2_i =  \bC_r(f^1_p,f^1_n,\bx_i^1,y) \\
&\tilde \delta^2_i=\delta^3+ \bC(yf^3(\bx_i^3))
\end{align}
Minimize over $f_3$ and keep $f_p^1,f^1_n,f_p^2,f_n^2$ constant:
\begin{align}
&\arg \min_{f^3} \sum_{i} R(f^1_p,f^1_n,f^2_p,f^2_n,f^3,\bx_i,y_i)= \\
&\arg \min_{f^3} \sum_{i}S^3_i  \bC(yf^3(\bx_i^3) \\
&\mbox{such that:} \\
&S^3_i =  \bC_r(f^1_p,f^1_n,\bx_i^1,y) \bC_r(f^2_p,f^2_n,\bx_i^2,y)
\end{align}


\end{document}